\documentclass[acmtog,prologue,table,xcolor]{acmart}

\usepackage[font=small, labelfont=bf]{caption}
\usepackage{makecell}
\usepackage{mathtools}
\usepackage{multirow}
\usepackage{subcaption}
\usepackage{algorithmic}
\usepackage{balance}
\usepackage{tcolorbox}

\usepackage{amsmath}
\usepackage{bm}
\usepackage[ruled,vlined,linesnumbered]{algorithm2e}
\usepackage{graphicx}
\usepackage[safe]{tipa}
\usepackage{multirow}
\usepackage{xcolor}
\usepackage{url}

\usepackage{pifont}
\makeatletter
\newcommand\HUGE{\@setfontsize\Huge{20}{30}}
\makeatother

\def\eqref#1{equation~\ref{#1}}

\def\1{\bm{1}}

\DeclareMathAlphabet{\mathsfit}{\encodingdefault}{\sfdefault}{m}{sl}
\SetMathAlphabet{\mathsfit}{bold}{\encodingdefault}{\sfdefault}{bx}{n}

\newcommand{\datasetName}{KnitCAD}

\definecolor{pink}{RGB}{255, 192, 203}
\definecolor{lightblue}{HTML}{0C8EF4}
\definecolor{lightorange}{HTML}{F87D3A}
\definecolor{violet}{HTML}{C299DD}
\definecolor{tealgreen}{HTML}{44BB99}
\definecolor{lightyellow}{HTML}{D0B866}
\definecolor{apricot}{HTML}{DD9977}
\definecolor{lightpurple}{HTML}{7D58D9}
\definecolor{lightgray}{HTML}{898989}

\AtBeginDocument{%
  }

\copyrightyear{2026}
\acmYear{2026}
\setcopyright{cc}
\setcctype{by}
\acmConference[SA Conference Papers '26]{SIGGRAPH Asia 2026 Conference Papers}{December 01--04, 2026}{Kuala Lumpur, Malaysia}
\acmBooktitle{SIGGRAPH Asia 2026 Conference Papers (SA Conference Papers '26), December 01--04, 2026, Kuala Lumpur, Malaysia}
\acmDOI{10.1145/3829340.3842261}
\acmISBN{979-x-xxxx-xxxx-x/YYYY/MM}

\begin{document}

\title{\textit{CADKnitter}: Compositional CAD Generation from Text and Geometry Guidance}

\author{Tri Le}
\affiliation{%
  \institution{Quantum AI \& Cyber Security Institute, FPT Corporation}
  \country{Vietnam}
}
\email{trilq3@fpt.com}

\author{Khang Nguyen}
\affiliation{%
  \institution{Department of Robotics, Mohamed bin Zayed University of Artificial Intelligence}
  \country{UAE}
}
\email{khang.nguyen@mbzuai.ac.ae}

\author{Baoru Huang}
\affiliation{%
  \institution{Department of Computer Science, University of Liverpool}
  \country{UK}
}
\email{baoru.huang@liverpool.ac.uk}

\author{Tung D. Ta}
\affiliation{%
 \institution{Faculty of Environment and Information Studies, Keio University}
 \country{Japan}
}
\email{tung@keio.jp}

\author{Anh Nguyen}
\affiliation{%
  \institution{Department of Computer Science, University of Liverpool}
  \country{UK}
}
\email{anh.nguyen@liverpool.ac.uk}

\renewcommand{\shortauthors}{Le et al.}

\begin{abstract}
  Computer-aided design (CAD) defines 3D models as compact, precise, and editable representations, making it directly useful for several fields. Recently, CAD generation has been gaining more attention in both the research community and industry. Crafting CAD models has long been a painstaking and time-intensive task, demanding both precision and expertise from designers.
  Prior works have achieved early success in single-part CAD generation, which is not well-suited for real-world applications, as multiple parts need to be assembled under semantic constraints and geometric compatibility. In this paper, we propose \textit{CADKnitter}, a compositional CAD generation framework with geometric-guiding cues to steer diffusion sampling. \textit{CADKnitter} is able to generate a complementary CAD part that follows both the geometric constraints of the given CAD model and the semantic constraints of the desired design text prompt. We also curate a dataset, so-called \textit{KnitCAD}, containing over $310,000$ samples of CAD models, along with textual prompts and assembly metadata that provide semantic and geometric constraints. Intensive experiments demonstrate that our proposed method outperforms other state-of-the-art baselines by a clear margin. Our project page is available at \href{https://cadknitter.github.io/}{https://cadknitter.github.io}.
\end{abstract}

\ccsdesc[500]{Computing methodologies~Computer vision}

\keywords{CAD Model, B-rep representation, Compositional CAD Generation}

\begin{teaserfigure}
  \includegraphics[width=\textwidth]{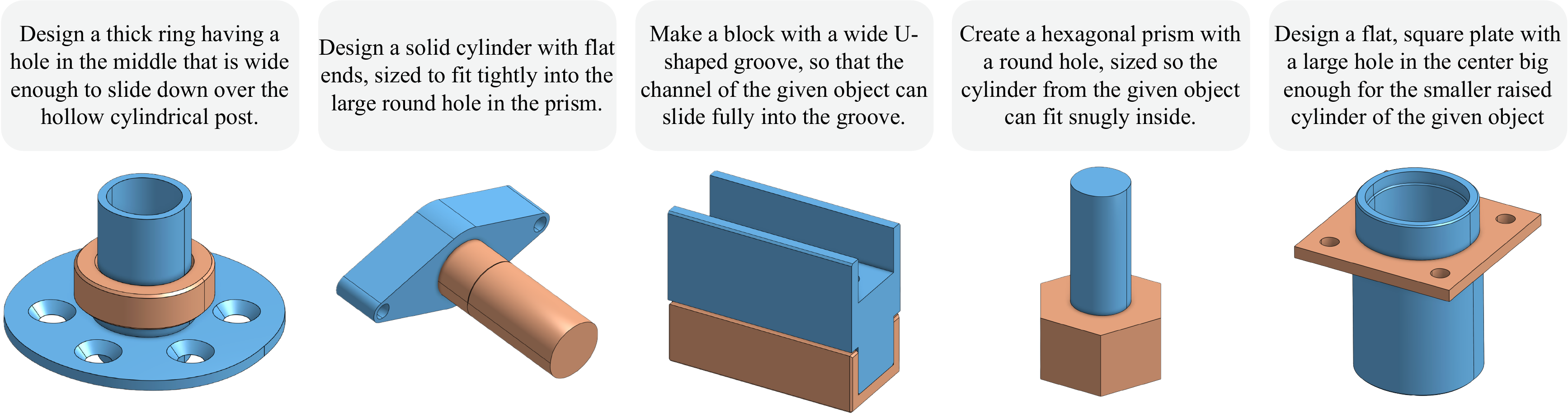}
  \caption{\textbf{The compositional CAD generation task.} Our \textit{CADKnitter} model takes in a \textcolor{lightgray}{\textbf{text prompt}} and a \textcolor{lightblue}{\textbf{conditioned CAD model}} to generate a new \textcolor{lightorange}{\textbf{complementary CAD model}} that geometrically fits with the input CAD and semantically aligns with the design text prompt.}
  \Description{\textbf{The compositional CAD generation task.} Our \textit{CADKnitter} model takes in a \textcolor{lightgray}{\textbf{text prompt}} and a \textcolor{lightblue}{\textbf{conditioned CAD model}} to generate a new \textcolor{lightorange}{\textbf{complementary CAD model}} that geometrically fits with the input CAD and semantically aligns with the design text prompt.}
  \label{fig:teaser}
\end{teaserfigure}

\maketitle

\section{Introduction}
\label{sec:introduction}
Computer-aided design (CAD) generation has been gaining attention in research and industry~\cite{li2025cad, li2025caddreamer}. Since CAD models are precise and editable parametric representations, generating CAD models has several applications in real-world design and manufacturing tasks such as product design, mechanical engineering~\cite{makatura2023can}, and robotics~\cite{narang2022factory, wang2025matchmaker, tian2025fabrica}. In practice, designing CAD models remains a challenging and labor-intensive process across industrial and automated manufacturing domains as it requires both precision and expertise from designers. To improve productivity and reduce manual effort, recent research has explored automated frameworks for CAD generation~\cite{khan2024text2cad, wang2025text, li2025cad, xu2024cad}.

Recent efforts have shown promising results in unconditional CAD generation~\cite{xu2024brepgen, lee2025brepdiff, xu2025autobrep}, text-conditioned generation~\cite{khan2024text2cad, wang2025text}, image-based reconstruction~\cite{li2025caddreamer, xu2024cad}, and point cloud-based rendering~\cite{liu2024point2cad}. Despite these advancements, existing generation methods primarily focus on \textit{single CAD generation} and overlook the complex inter-dependencies between multiple parts. In practice, CAD not only requires a single model; it typically involves an assembly of multiple parts subjected to strict geometric and semantic constraints~\cite{willis2021joinable, jones2021automate, tian2022assemble}. Moreover, the current state of CAD generation is inapplicable to real assemblies due to its object-centric nature, lack of assembly awareness, and geometry-guided mechanisms for enforcing how such components should be aligned together.

Practically, CAD demands not only the creation of individual parts but also meticulous precision in how distinct parts align and assemble~\cite{jones2021automate, willis2021joinable}. 
To support this need, we introduce a new practical problem for the community: compositional CAD generation. This problem aims to generate CAD models with awareness of existing CAD models while considering both semantic and geometric constraints, as shown in Fig.~\ref{fig:teaser}. Semantic constraints refer to the functional and contextual relationships between parts, ensuring that the components align with the user-intended design (\textit{e.g.}, ``generating a long cylindrical shaft to fit through a hole''). Meanwhile, geometric compliance defines the spatial and structural relationships that govern how the generated CAD parts fit and align with the existing ones. While single CAD generation inherently struggles with the internal structural geometry to ensure output validity~\cite{xu2024brepgen}, compositional CAD generation introduces additional geometric compatibility with external CAD parts.

In this paper, we propose \textit{CADKnitter}, a method for enhancing the CAD design process. 
\textit{CADKnitter} firstly introduces a compositional CAD generation diffusion model. Building upon prior text-to-CAD approaches that focus solely on single-object fidelity, \textit{CADKnitter} explicitly models inter-part relationships, enabling it to synthesize components that fit together. Furthermore, based on our statistical analysis, we propose a geometric-guiding cue to explicitly enforces geometric compatibility during sampling process by optimizing contact faces between the generated and conditional parts. 
We further construct \textit{KnitCAD}, a large-scale dataset comprising text–CAD pairs with detailed assembly metadata that enables scalable learning under semantic–geometric constraints. The intensive experiments show that our method outperforms recent baselines. In summary, our contributions are threefold:
\begin{itemize}
    \item We introduce \textit{KnitCAD}, a large-scale dataset for the compositional CAD generation of over $310,000$ text-CAD pairs with detailed assembly metadata and automatically annotated contact faces.
    \item We propose a compositional CAD generation network and a geometric-guiding cue that is used to enforce geometric compatibility, while preserving semantic alignment.
    \item We demonstrate that our methods significantly improve geometric compatibility and semantic fidelity compared to state-of-the-art baselines.
\end{itemize}

\begin{figure*}[ht]
    \centering
    \includegraphics[width=\linewidth]{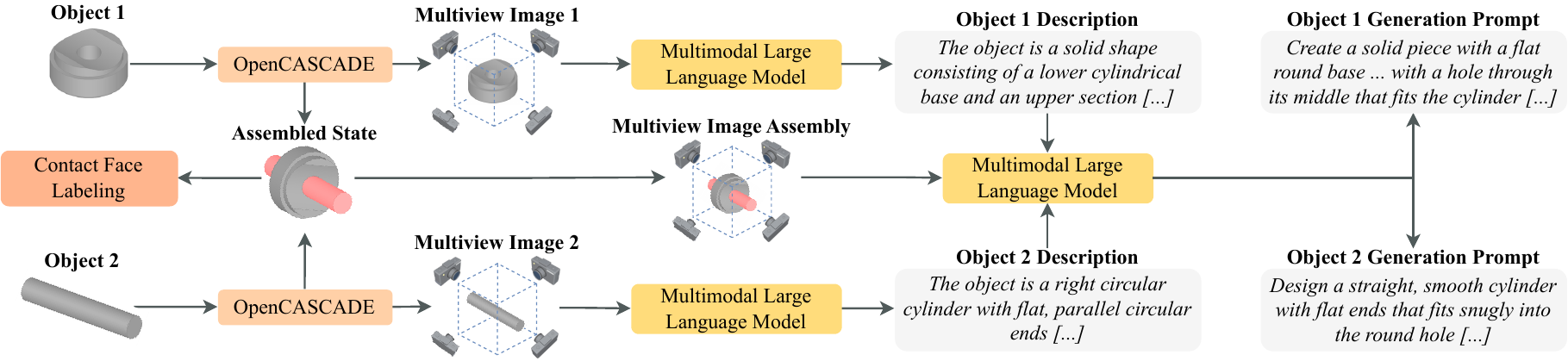}
    \caption{\textbf{Construction of \textit{KnitCAD}.} The dataset is curated from prior CAD datasets. For each pair of B-rep models, OpenCASCADE~\cite{opencascade} is used to render multiview images and label contact faces in their fully assembled states. Multimodal LLM generates textual descriptions of shapes and prompts for generation from rendered images.}
    \label{fig:dataset-construction}
\end{figure*}
\section{Related Work}
\subsection{CAD Generation} 
Many methods have been proposed for CAD generation, such as text to CAD~\cite{wang2025text, khan2024text2cad, xu2024cad}, image to CAD~\cite{chen2025cadcrafter, li2025caddreamer, you2024img2cad}, or point cloud CAD rendering~\cite{liu2024point2cad, guo2022complexgen}. Among these works, several methods represent a CAD model as a sequence of sketch and extrusion operations, generating it in an auto-regressive fashion~\cite{wu2021deepcad, khan2024text2cad, xu2022skexgen, wang2025text, zhang2024flexcad}. Other work~\cite{yu2022capri, ren2021csg, mews2024don} focuses on generating Constructive Solid Geometry, which represents 3D shapes through hierarchies of Boolean operations and basic primitive shapes. While these representations are flexible for generative models, their capability to present complex CAD models is limited~\cite{jayaraman2022solidgen, li2024sfmcad, yu2023d}. Instead, the predominant format for CAD models is Boundary-representation (B-rep)~\cite{ansaldi1985geometric, weiler1986topological, stroud2011solid, lee2025brepdiff}. B-rep presents a CAD model by combining the continuous geometry and discrete topology of primitives, which is challenging for current generative models to learn for both data types. To synthesize B-rep models directly, several works~\cite{jayaraman2022solidgen, xu2024brepgen, li2025dtgbrepgen} propose cascaded generative models, while others~\cite{guo2025brepgiff, lee2025brepdiff, liu2025hola} explore unified representations that encode both types of data. In this work, we use B-rep representation due to its ability to model real-world CAD models.

\subsection{Compositional Shape Generation} 
Recent research has investigated the synthesis of object parts through different representation inputs, such as images and point clouds~\cite{chen2025partgen, yan2024phycage, wang2025matchmaker, yang2025omnipart, liu2024singapo, luo2025physpart, liu2024cage}. For example, OmniPart~\cite{yang2025omnipart} constructs separate 3D parts from single images. SINGAPO~\cite{liu2024singapo} proposes a retrieval-based approach to construct articulated objects from images. Recently, MatchMaker~\cite{wang2025matchmaker} is a multi-stage framework that produces parts satisfying stability constraints from given parts for robotic simulation. These works aim to output final objects in mesh/point cloud format and lack experiments on geometric compatibility which is essential for 3D CAD modeling. In general, prior works typically focus on the semantic relationship and overlook the fine-grained geometric constraints between distinct parts.

In this paper, we address compositional CAD generation, guided by both semantic descriptions from input text prompts and geometric compatibility constraints from provided CAD models that adhere to real-world 3D CAD design process. Also, we focus on modeling 3D CAD parts at their 3D printing-ready theoretical size (nominal size) because these output CAD can be used or post-processed differently for a wide range of downstream applications such as simulation, fabrication, and manufacturing.

\subsection{Diffusion Guided Sampling} 
Guided sampling strategies for diffusion models can be broadly categorized into gradient-based~\cite{wang2023diffusebot, nguyen2024language, yang2024physcene, nguyen2025egomusic, chen2024atlas3d, du2023reduce} and derivative-free~\cite{li2024derivative, yuan2023physdiff, xu2023interdiff, flaxman2004online, guo2025training} mechanisms. These approaches enforce specific constraints during the generation process. For instance, PhysDiff~\cite{yuan2023physdiff} employs a motion projection module to adjust intermediate steps in the reverse diffusion process, while SVDD~\cite{li2024derivative} and Zero-Order Search~\cite{flaxman2004online} frame sampling as a search-based optimization problem to steer generation. These derivative-free strategies are particularly well-suited for non-differentiable pipelines, often requiring a scoring function to assess intermediate states. In this work, we adopt this guidance framework and introduce a scoring function formulated around our proposed geometric-guiding cues to search for the best sampling trajectory.

\section{The \textit{\datasetName{}} Dataset}
\label{sec:dataset}

\begin{table}[hbt]
    \centering
    \caption{\textbf{Comparison of recent CAD generation datasets.} Our \textit{KnitCAD} dataset enables the compositional CAD generation task by providing both the contact face label, text instruction, and the conditioned CAD model.} \label{table:assembly_dataset}
    \renewcommand{\tabcolsep}{5pt}
    \resizebox{\linewidth}{!}{
        \begin{tabular}{l|ccccc}
        \toprule
        \makecell{\textbf{Dataset}} & \makecell{\textbf{No.} \\ \textbf{Samples}} & \makecell{\textbf{Repre-} \\ \textbf{sentation}} & \makecell{\textbf{Object Pair} \\ \textbf{Label}} & \makecell{\textbf{Contact Face} \\ \textbf{Label}} & \makecell{\textbf{Text}} \\ 
        \midrule
        ABC~\cite{koch2018abc} & 1M & B-rep & \textcolor{red!40}{\ding{56}} & \textcolor{red!40}{\ding{56}} & \textcolor{red!40}{\ding{56}} \\
        DeepCAD~\cite{wu2021deepcad} & 178K & B-rep & \textcolor{red!40}{\ding{56}} & \textcolor{red!40}{\ding{56}} & \textcolor{red!40}{\ding{56}} \\
        Fusion 360 Joint~\cite{willis2021joinable} & 19K & B-rep & \textcolor{green}{\ding{52}} & \textcolor{green}{\ding{52}} & \textcolor{red!40}{\ding{56}} \\
        Zero-to-CAD~\cite{ataei2026zero} & 1M & B-rep & \textcolor{red!40}{\ding{56}} & \textcolor{red!40}{\ding{56}} & \textcolor{red!40}{\ding{56}} \\
        AutoMate~\cite{jones2021automate} & 542K & B-rep & \textcolor{green}{\ding{52}} & \textcolor{red!40}{\ding{56}} & \textcolor{red!40}{\ding{56}} \\
        2BY2~\cite{qi2025two} & 517 & Mesh & \textcolor{green}{\ding{52}} & \textcolor{red!40}{\ding{56}} & \textcolor{red!40}{\ding{56}} \\
        ATA~\cite{tian2022assemble} & 8800 & Mesh & \textcolor{green}{\ding{52}} & \textcolor{red!40}{\ding{56}} & \textcolor{red!40}{\ding{56}} \\
        Factory~\cite{narang2022factory} & 60 & B-rep & \textcolor{green}{\ding{52}} & \textcolor{red!40}{\ding{56}} & \textcolor{red!40}{\ding{56}} \\ 
        Text2CAD~\cite{khan2024text2cad} & 170K & B-rep & \textcolor{red!40}{\ding{56}} & \textcolor{red!40}{\ding{56}} & \textcolor{green}{\ding{52}} \\
        CADFusion~\cite{wang2025text} & 20K & B-rep & \textcolor{red!40}{\ding{56}} & \textcolor{red!40}{\ding{56}} & \textcolor{green}{\ding{52}} \\
        Omni-CAD~\cite{xu2024cad} & 453K & B-rep & \textcolor{red!40}{\ding{56}} & \textcolor{red!40}{\ding{56}} & \textcolor{green}{\ding{52}} \\
        \midrule
        \rowcolor[HTML]{FFF2CC} 
        \textbf{\textit{KnitCAD} (Ours)} & \textbf{157K} & \textbf{B-rep} & \textcolor{green}{\ding{52}} & \textcolor{green}{\ding{52}} & \textcolor{green}{\ding{52}} \\ \bottomrule
        \end{tabular}
    }
\end{table}

Our proposed \textit{\datasetName{}} dataset builds upon two recent datasets: (\textit{i}) Fusion 360 Gallery Assembly – Joint Data (Fusion 360 Joint)~\cite{willis2021joinable} contains $19,156$ joint pairs from $23,029$ B-rep models, and (\textit{ii}) AutoMate~\cite{jones2021automate} includes $541,635$ mate pairs from $376,362$ B-rep models. These datasets provide detailed joint-axis annotations that fully describe the connection between two CAD parts, making them directly relevant to our compositional CAD generation task under geometric constraints and text descriptions.

Specifically, we extend the original datasets~\cite{willis2021joinable,jones2021automate} by generating text descriptions for each pair using state-of-the-art multi-modal large language models (MLLM) and by labeling the contact faces of every B-rep model with OpenCASCADE~\cite{opencascade}. Here, a contact face (\textit{i.e.}, a bounded surface in B-rep representation) is defined as a face that is within a small distance tolerance of a face on the complementary CAD model. These contact-face labels provide important geometric information, indicating exactly how the parts interact in the assembled state. The comparisons among the attributes of open-sourced CAD datasets and our curated \textit{KnitCAD} dataset are illustrated in Table~\ref{table:assembly_dataset}.

\subsection{Generating Prompts for CAD Generation}
For each pair of B-rep models, we synthesize two generation prompts that describe the target CAD model and how they are semantically complementary to each other, as shown in Fig.~\ref{fig:dataset-construction}. We first render multi-view images of each part and their assembled configuration using OpenCASCADE~\cite{opencascade}. These rendered views are then fed into an MLLM to produce textual descriptions for each object; meanwhile, another MLLM subsequently fuses the individual object descriptions to form paired assembly prompts that capture the semantic intent. Note that we employ GPT-4.1~\cite{openai2025chatgpt} as the main MLLM for all text generation. To control the quality generated by LLM, we manually verify a subset of 1,000 random samples to ensure the quality of generation before applying the automated generation pipeline to the remaining dataset.

For the Fusion 360 Joint dataset, we select one representative joint per model pair, while for AutoMate, duplicate mate definitions between identical model pairs are removed to ensure uniqueness. In total, we curate $156,654$ unique assembled pairs derived from $172,265$ distinct B-rep models. Therefore, our dataset contains $313,308$ samples, each of which includes a target CAD model, a condition CAD model, and a textual prompt for learning CAD generation.

\begin{figure*}[t]
    \centering
    \includegraphics[width=\linewidth]{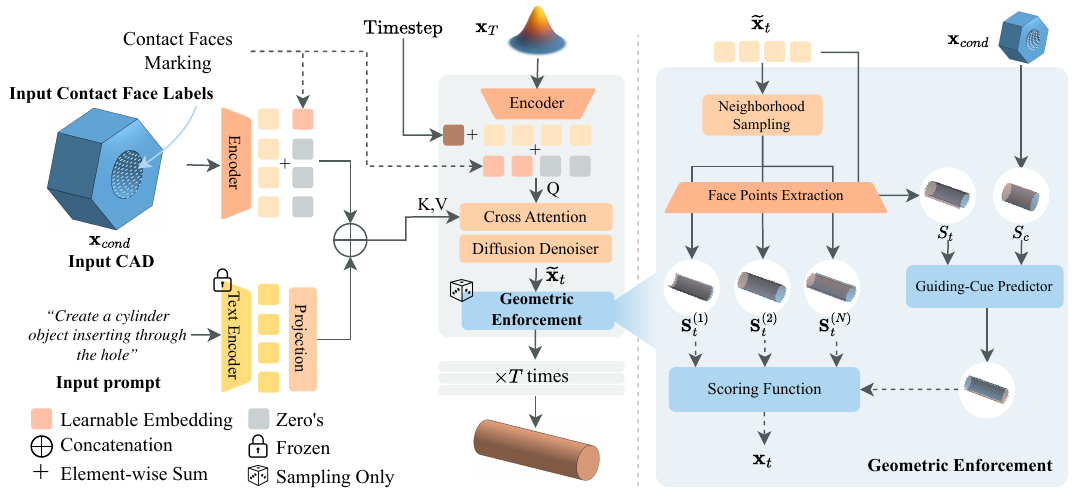}
    \caption{\textbf{Overview of \textit{CADKnitter}.} We propose a compositional CAD generation network \textit{(left)}. To enforce the geometric constraint, we propose a scoring function and geometric-guiding cues which are used in a derivative-free guidance mechanism to steer the generation \textit{(right)}.}
    \label{fig:method-overview}
\end{figure*}

\subsection{Contact Face Conditions and Labeling}
\label{sec:contact_face_labeling}
To enable compositional CAD generation, we use OpenCASCADE to label the contact faces between two CAD models. Two faces are defined to be in contact if they have common curvatures that overlap or are within a small tolerance of each other. We uniformly sample a discretized face $s = \{\mathbf{p}_{1}, \mathbf{p}_{2}, \dots, \mathbf{p}_{N_{s}} \}$ of $N_{s}$ points from a parametric surface, where each point $\mathbf{p}_{i} \in \mathbb{R}^{3}$, with $N_{s} = 32 \times 32$~\cite{jayaraman2021uv}. The conditions for overlapping or being within a small tolerance between a point $\mathbf{p} \in s$ and a face $s'$ are mathematically described by the following properties:
\begin{itemize}
    \item \underline{Point-to-Surface Proximity:} $\|\mathbf{p} - \mathbf{q}\|_2 \le \delta$, where $\mathbf{q} \in s'$ is defined as the closest point to $p$ and $\delta$ represents the tolerance threshold,
    \item \underline{Inverted Normals:} $\mathbf{n}(\mathbf{p}) \cdot \mathbf{n} \left(\text{Pr}_{s'}(\mathbf{p})\right) < 0 $, where $\text{Pr}_{s'}(\mathbf{p}) \in \mathbb{R}^{3}$ denotes the projection of $\mathbf{p}$ onto the parametric surface of $s'$ and $\mathbf{n}(\cdot) \in \mathbb{R}^{3}$ represents the face normal vector at the specified point.
\end{itemize}
Therefore, two faces $s_{1}$ and $s_{2}$ are said to be in contact if a subset of points $\mathbf{p}_{i, 1} \in s_{1}$ satisfies both conditions with respect to the face $s_{2}$, or vice versa. These serve as explicit conditions for compositional CAD generation in our dataset.
\section{Compositional CAD Generation}
\label{sec:methodology}

\subsection{Problem Formulation} 
In our work, we represent the CAD model as a B-rep model with a set of face entities $\mathbf{x} = \{x^{(i)}\}_{i=1}^{N}$, where $N$ is the number of faces, each element $x^{(i)} \in \mathbb{R}^{D_{x}}$ represents a B-rep face geometric encoding. It is noted that $D_{x}$ can be either bounding boxes~\cite{xu2024brepgen} or UV-sampled geometric points~\cite{jayaraman2021uv}. 
The edges, vertices, and the topological information are represented as other sets and adjacency matrices~\cite{li2025dtgbrepgen, xu2024brepgen}. 
For brevity, we omit the explicit formulation.

Given a text prompt $\mathcal{T}$, a B-rep model $\mathbf{x}_{\text{cond}}$, and a set of indices $\mathcal{I} \subseteq \{1, 2, \ldots, N\}$ for labeling the desired contact faces on $\mathbf{x}_{\text{cond}}$, compositional CAD generation aims to learn a model that can generate the complementary CAD $\mathbf{x}'$ satisfying both semantic and geometric constraints. We address the compositional CAD generation problem by using diffusion model~\cite{ho2020denoising, song2020denoising} with the target $\mathbf{x}'$ as $\mathbf{x}_0$, and the objective is to learn the reverse process $p_\theta(\mathbf{x}_{t-1}|\mathbf{x}_{t}, \mathbf{c})$ with condition $\mathbf{c}$. Sec.~\ref{sec:method-network} presents our proposed network. As in previous work~\cite{xu2024brepgen, jayaraman2022solidgen, li2025dtgbrepgen}, we use a diffusion-based CAD generation for generating B-rep entities sequentially, including face bounding boxes, face points, edge bounding boxes, and edge points. However, diffusion models often fail to strictly satisfy geometric constraints due to their stochastic nature and approximation errors in denoising~\cite{yuan2023physdiff}. Therefore, given the non-differentiable CAD generation pipeline, we employ a derivative-free guidance in the sampling process, in which we propose a new scoring function and geometric-guiding cues (Sec. \ref{sec:method-guidance}).

\subsection{CADKnitter Overview}
\label{sec:method-network}
The overall pipeline of our proposed network is illustrated in Fig.~\ref{fig:method-overview} (left). For semantic conditioning, we encode the text prompt as embeddings $\mathbf{e}_\tau$ using a pre-trained text encoder followed by a linear projection. For geometric conditioning, each element in the conditional B-rep is embedded as $\mathbf{e}_c$ using a dedicated encoder. To emphasize contact face signal, we inject learnable embeddings at corresponding elements in the conditional input. Similarly, for the noisy sample $\mathbf{x}_t$, we predefine a set of $M$ elements as contact faces and also inject the shared, learnable tokens. Finally, the two condition signals are concatenated into a single sequence $\mathbf{c}$ and fused with the noisy representation $\mathbf{x}_t$ via a cross-attention mechanism before processing with a transformer-based denoiser. 

\subsection{Geometric Enforcement}
\label{sec:method-guidance}
Derivative-free guidance frames the sampling process as a search-based optimization problem requiring a scoring function to evaluate intermediate samples. Within this framework, our core contribution is a scoring function centered on our proposed geometric-guiding cues. In principle, these guiding cues provide the potential geometry of current intermediate samples' contact faces which align with the condition faces. By leveraging these cues, the scoring function evaluates candidate samples to guide the sampling path.

\subsubsection{Derivative-free guidance.} The overall guidance mechanism is illustrated in Fig.~\ref{fig:method-overview} (right). Specifically, we consider Zero-Order Search~\cite{flaxman2004online} as a guidance mechanism to enforce the geometric constraint:
\begin{enumerate}
    \item Sample denoised output $\tilde{\mathbf{x}}_t \sim p_\theta(\tilde{\mathbf{x}}_{t}\mid \mathbf{x}_{t+1}, \mathbf{c})$.
    \item Generate $N_p$ candidates in the neighborhood: $\{\hat{\mathbf{x}}_t^{(n)}\}_{n=1}^{N_p} \sim p_\theta(\hat{\mathbf{x}}_t \mid \tilde{\mathbf{x}}_t, \mathbf{c})$.
    \item Get face points $\{\mathbf{S}_t^{(n)}\}$ from $\{\hat{\mathbf{x}}_t^{(n)}\}$ and $\tilde{\mathbf{S}}_t$  from $\tilde{\mathbf{x}}_t$.
    \item Compute candidate scores $\{R(\mathbf{S}_t^{(n)})\}$ and select $\hat{\mathbf{x}}_t^{(n)}$ with the best scores as next intermediate samples.
\end{enumerate}
The neighborhood distribution $p_\theta(\hat{\mathbf{x}}_t \mid \tilde{\mathbf{x}}_t, \mathbf{c})$ in step \textit{(2)} admits several definitions~\cite{ma2025inference, li2024derivative}; a common choice is an isotropic Gaussian centered on the denoised sample $\tilde{\mathbf{x}}_t$. To then score these candidates, face points can be extracted from intermediate samples $\mathbf{x}_t$ via specific diffusion models. This technique enables the evaluation of intermediate steps without constructing a final, valid CAD model, thereby significantly reducing computational overhead. To ensure accuracy, we apply this guidance during the later stages of the reverse process, where $\mathbf{x}_t$ exhibits lower noise levels and the extracted points are more reliable.

\subsubsection{Scoring Function.} We propose a scoring function for a face point set $\mathbf{S}_t^{(n)}$ as a weighted sum of geometry alignment term $D_{\text{geo}}$ and semantic preservation term $D_{\text{sem}}$:
\begin{equation}
    R\bigl(\mathbf{S}_t^{(n)}\bigr) = D_{\mathrm{geo}} \bigl(\mathbf{S}_t^{(n)}\bigr) + \lambda_\text{sem}D_{\text{sem}} \big( \mathbf{S}_t^{(n)} \big).
\end{equation}

\textit{Geometry Alignment $D_{\text{geo}}$}. The geometry alignment term penalizes geometric discrepancies between the \textit{contact faces} of a candidate sample and the faces of a \textit{guiding cue} obtained from our proposed predictor (see Sec.~\ref{sec:guiding-sample-predictor}).
Let $S_1 \subset \mathbf{S}_t^{(n)}$ and $S_2$ denote point sets of the candidate's contact faces and the guiding cue's faces, respectively. The term $D_{\text{geo}}$ is formulated using the Chamfer distance:

\begin{equation}
\begin{split}
    D_{\mathrm{geo}} \bigl(\mathbf{S}_t^{(n)}\bigr) = 
    &\frac{1}{|S_1|}\sum_{x \in S_1} \min_{y \in S_2} \|x - y\|_2^2  + \frac{1}{|S_2|}\sum_{y \in S_2} \min_{x \in S_1} \|x - y\|_2^2.
\end{split}
\end{equation}

\textit{Semantic Preservation $D_{\text{sem}}$}. While the geometry alignment term encourages the local fitness of contact faces, the global structure must be maintained to preserve the semantic meaning of the original intermediate sample. We employ the Fused Gromov–Wasserstein distance~\cite{vayer2018optimal} to quantify the scale-invariant structural similarity between the candidate sample $\mathbf{S}^{(n)}_t$ and the original intermediate sample $\tilde{\mathbf{S}}_t$. We utilize the bounding boxes of the point sets for each face to represent this global structural information. Specifically, we use the bounding box aspect ratio $r_{i} \in \mathbb{R}^3$ as a feature of each face $s_i$ in $\mathbf{S}_t^{(n)}$ and $\tilde{\mathbf{S}}_t$, while measuring scale-invariant distances between normalized bounding box centers. Let $\mathbf{C},\mathbf{C}'$ and $r, r'$ denote the normalized centers and aspect ratios of the bounding boxes in the two sets, respectively. The term $D_{\text{sem}}$ is given by:
\begin{equation}
    \begin{aligned}
        D_{\text{sem}} \big( \mathbf{S}_t^{(n)} \big) &= (1-\lambda) \sum_{i,i',j,j'} W_{i,i',j,j'}^{2} T_{ij}T_{i'j'} + \lambda \sum_{i,j} \|r_{i} - r'_{j}\|_{2}^{2} \, T_{ij} \\
        W_{i,i',j,j'} &= \|\mathbf{C}_{i} - \mathbf{C}_{i'}\|_{2} - \|\mathbf{C}'_{j} -\mathbf{C}'_{j'}\|_{2},
    \end{aligned}
    \label{eq:fgw} 
\end{equation}
where $\lambda$ is predefined and $T$ represents the transport plan. We use Sinkhorn algorithm to solve the transport plan $T$.

\subsubsection{Guiding-Cue Predictor}
\label{sec:guiding-sample-predictor}  
We obtain the geometric-guiding cue by optimizing the contact faces of intermediate samples to geometrically align with the conditional contact faces while maintaining their original semantic structure.
Formally, the guiding-cue predictor returns a set of optimized contact faces from the intermediate samples.
We denote ${S}_{t}$ and ${S}_{c}$ as point sets of contact faces on the intermediate samples and the condition CAD model, respectively.
We optimize the translation and shape of $S_{t}$ to align with $S_{c}$ by minimizing the following cost function:
\begin{equation}
    \mathcal{C} = \mathcal{C}_{\text{pos}} + \mathcal{C}_{\text{shape}},
    \label{eq:total_cost}
\end{equation}
where $\mathcal{C}_{\text{pos}}$ is the positional cost function and $\mathcal{C}_{\text{shape}}$ denotes the shape cost function. We design these cost functions to prevent the optimized contact faces from collapsing into the exact shape of the condition.

\textit{Positional cost:} This cost encourages generated contact faces $S_{t}$ to minimize its distance to the corresponding conditional face $S_{c}$. Formally:
\begin{equation}
    \mathcal{C}_{\text{pos}} \left(S_{t}, S_{c}\right) = \min_{p \in S_{t}} \min_{\tau \in \text{T}\left(S_{c}\right)} d({p, \tau}),
    \label{eq:pos_cost}
\end{equation}
where $d(p, \tau)$ denotes the point-to-mesh distance function from a point $p$ on $S_{t}$ to the triangle $\tau$ tessellated from the triangle sets $\text{T}(\cdot)$ of $S_{c}$. Minimizing $\mathcal{C}_{\text{pos}}$ ensures the new contact faces have a region in contact with the conditioned faces.

\begin{figure}[h]
    \centering
    \includegraphics[width=\linewidth]{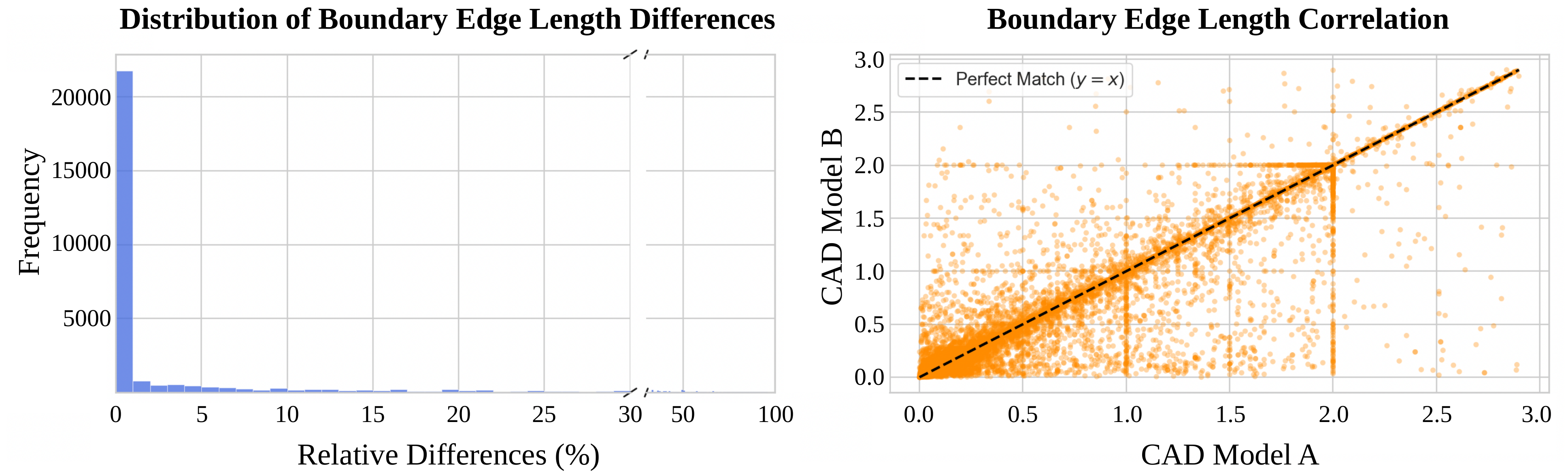}
    \caption{\textbf{Statistical analysis on boundary edge length of contact faces.} \textit{(Left)} Distribution of relative length differences between paired contact boundary edges. \textit{(Right)} Scatter plot of boundary edge lengths between two mating parts.}
    \label{fig:edge-length-analysis}
\end{figure}

\begin{table*}[h]
    \centering
    \begin{minipage}[b]{0.32\linewidth}
        \caption{\textbf{Main results.}}
        \label{table:compositional-text2cad-generation}
        \centering
        \setlength{\tabcolsep}{3.5pt} 
        \resizebox{\linewidth}{!}{
            \begin{tabular}{r | cccc}
                \toprule
                \textbf{Method} & $\textbf{CD} \downarrow$ & $\textbf{PR} \downarrow$ & $\textbf{IV} \downarrow$ & $\textbf{VR} \uparrow$ \\ 
                \midrule
                PivotMesh~\cite{weng2024pivotmesh} & 137.70 & - & - & - \\
                BrepGen~\cite{xu2024brepgen} & 116.82 & 0.43 & 16.71 & \textbf{0.47} \\
                MatchMaker~\cite{wang2025matchmaker} & 102.15 & 0.42 & 18.95 & 0.29 \\
                \midrule
                Ours (w.o. guidance) & 88.69 & 0.24 & 9.42 & 0.45 \\
                \rowcolor[HTML]{FFF2CC} Ours & \textbf{83.38} & \textbf{0.23} & \textbf{7.59} & 0.44 \\
                \bottomrule
            \end{tabular}
        }
    \end{minipage}
    \hfill
    \begin{minipage}[b]{0.32\linewidth}
        \caption{\textbf{Human evaluations.}}
        \label{table:user-preference-study}
        \centering
        \setlength{\tabcolsep}{3.5pt}
        \resizebox{\linewidth}{!}{
            \begin{tabular}{l | cc}
                \toprule
                 & \textbf{Semantics} & \textbf{Geo. Compatibility} \\ 
                 \midrule
                \rowcolor[HTML]{FFF2CC} Win vs. PivotMesh & \textbf{0.84} & \textbf{0.86} \\
                Tie vs. PivotMesh & 0.11 & 0.09 \\ \midrule
                \rowcolor[HTML]{FFF2CC} Win vs. BrepGen & \textbf{0.56} & \textbf{0.70} \\
                Tie vs. BrepGen & 0.31 & 0.19 \\ \midrule
                \rowcolor[HTML]{FFF2CC} Win vs. MatchMaker & \textbf{0.49} & \textbf{0.67} \\
                Tie vs. MatchMaker & 0.37 & 0.19 \\  
                \bottomrule
            \end{tabular}
        }
    \end{minipage}
    \hfill
    \begin{minipage}[b]{0.32\linewidth}
        \caption{\textbf{Cross-set evaluation.}}
        \label{table:cross-set-eval}
        \centering
        \setlength{\tabcolsep}{3.5pt} 
        \resizebox{\linewidth}{!}{
            \begin{tabular}{r | cccc}
                \toprule
                \textbf{Method} & $\textbf{CD} \downarrow$ & $\textbf{PR} \downarrow$ & $\textbf{IV} \downarrow$ & $\textbf{VR} \uparrow$ \\ 
                \midrule
                PivotMesh~\cite{weng2024pivotmesh} & 142.56 & - & - & - \\
                BrepGen~\cite{xu2024brepgen} & 107.62 & 0.40 & 22.86 & \textbf{0.48} \\
                MatchMaker~\cite{wang2025matchmaker} & 106.44 & 0.51 & 23.40 & 0.10 \\
                \midrule
                Ours (w.o. guidance) & 84.61 & \textbf{0.26} & 13.91 & 0.47 \\
                \rowcolor[HTML]{FFF2CC} Ours & \textbf{82.00} & \textbf{0.26} & \textbf{13.90} & \textbf{0.48} \\
                \bottomrule
            \end{tabular}
        }
    \end{minipage}
\end{table*}

\textit{Shape cost:} While the positional cost globally translates the generated contact faces to align with the conditional faces, the shape cost locally optimizes their geometry. The generated contact faces must tightly fit the conditional faces while preserving structural variations. We first observe that assembled parts frequently exhibit insertion relationships. Specifically, we examine the boundary edges of their contact faces, which we define as edges not shared by multiple faces. The lengths of these edges are often equal or proportional. Indeed, Willis \textit{et al.}~\cite{willis2021joinable} found that this equality acts as a strong feature to predict the joint axes of two CAD parts.
We further statistically validate this observation on the AutoMate subset of KnitCAD, which is collected from user-uploaded CAD assemblies~\cite{jones2021automate} and thus reflects real-world design statistics. As shown in Fig.~\ref{fig:edge-length-analysis} (left), the distribution of relative length differences between paired contact boundary edges reveals that 75.25\% of assembled pairs have a relative difference of less than 5\%. Fig.~\ref{fig:edge-length-analysis} (right) presents the scatter plot of boundary edge lengths between two mating parts, where the points cluster closely along the perfect-match diagonal ($y = x$), yielding a Pearson correlation of 0.9480. To ensure this property is not dataset-specific, we further test its generalization on an independent CAD source in Sec.~\ref{sec:generalization-analysis}.

Driven by this strong statistical prior, we leverage boundary edge equality as the primary optimization objective for shape alignment. We formulate our shape cost function based on the boundary edges as follows: 
\begin{equation}
    \mathcal{C}_{\text{shape}} = \mathcal{C}_{\text{len}} + \lambda_{\text{angle}} \, \mathcal{C}_{\text{angle}},
    \label{eq:shape_cost}
\end{equation}

Let $e = \{e_i\}_{i=1}^{N_e}$ and $e' = \{e'_i\}_{i=1}^{N_e}$ denote the sequences of sampled boundary-edge points on the generated contact face $S_t$ and the conditional face $S_c$, respectively. The edge length cost $\mathcal{C}_{\text{len}}$ penalizes discrepancies between corresponding edge lengths:
\begin{equation*}
    \mathcal{C}_{\text{len}}(e, e') = \frac{1}{N_{e} - 1} \sum_{i=1}^{N_{e} - 1} (\| e_{i+1} - e_i \|_{2} - \| e'_{i+1} - e'_{i} \|_2)^{2}.
    \label{eq:edge_length_cost}
\end{equation*}
Meanwhile, the cost function for edge angles is computed as:
\begin{equation*}
    \mathcal{C}_{\text{angle}}(e, e') = \frac{1}{N_{e} - 1} \sum_{i=1}^{N_{e} - 1} \big(1 - \mathbf{u}_i \cdot \mathbf{u}'_i \big),
    \label{eq:edge_angle_cost}
\end{equation*}
where $\mathbf{u}_i = (e_{i+1} - e_{i}) / \| e_{i+1} - e_{i} \|_{2}$ and $\mathbf{u}'_{i} = (e'_{i+1} - e'_{i}) / \| e'_{i+1} - e'_{i} \|_{2}$ denote the unit direction vectors of the $i$-th edge segments on $S_{t}$ and $S_{c}$, respectively. 

\begin{figure}[ht]
    \centering

    \begin{subfigure}{\linewidth}
        \centering
        \includegraphics[width=\linewidth]{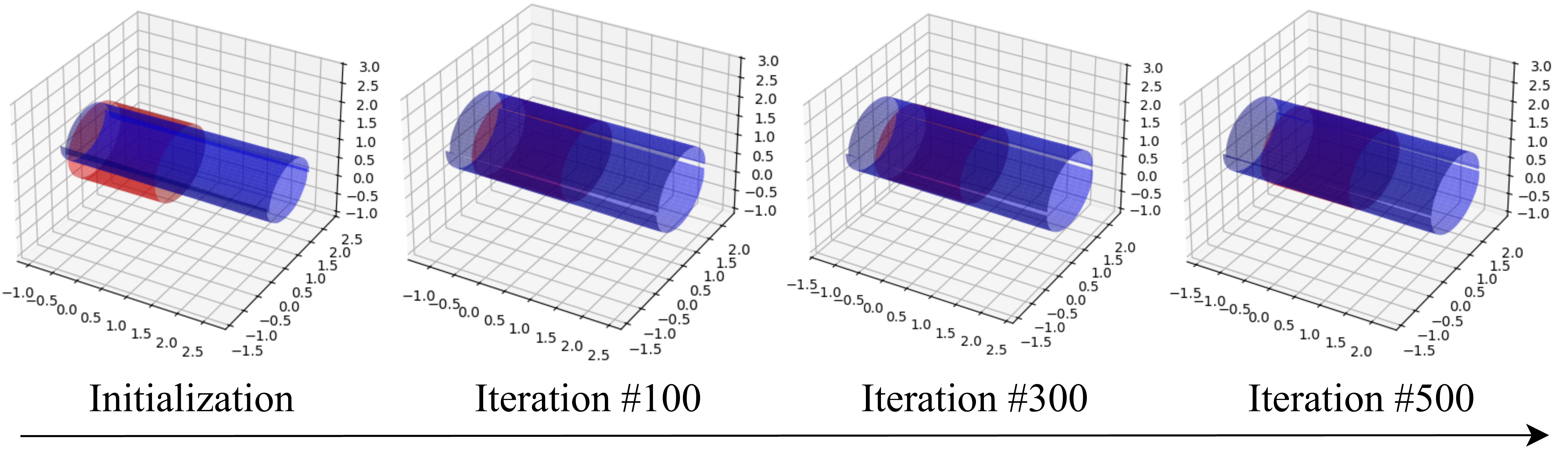}
        \caption{\textbf{Guiding-cue prediction.}}
        \label{fig:guiding-sample}
    \end{subfigure}


    \begin{subfigure}{\linewidth}
        \centering
        \includegraphics[width=\linewidth]{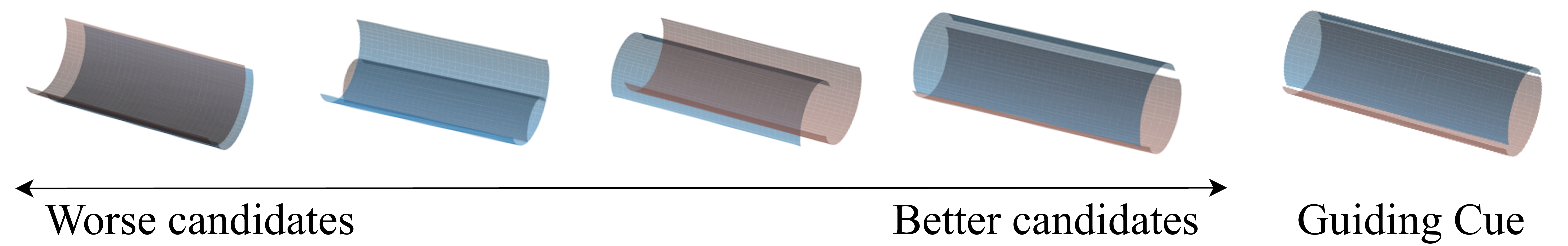}
        \caption{\textbf{Candidate scoring.}}
        \label{fig:candidates}
    \end{subfigure}

    \caption{\textbf{Guiding-cue prediction and candidate alignment.}
    (a) An intermediate sample's contact faces (\textcolor{blue}{blue}) are iteratively optimized to align with the condition faces (\textcolor{red}{red}), yielding the geometric-guiding cue.
    (b) Candidate samples, ordered from worse to better geometric alignment with the predicted guiding cue.}
    \label{fig:guiding}
\end{figure}

To preserve the internal structure during shape optimization, each interior point is updated by the same row-wise and column-wise displacements as its corresponding boundary edge points, so the entire face deforms coherently as the boundary aligns to the condition. Fig.~\ref{fig:guiding-sample} illustrates how contact faces of an intermediate sample are iteratively optimized into a geometric-guiding cue, and Fig.~\ref{fig:candidates} demonstrates how it is used to score candidates.
More details are provided in the Supplemental Material.

\section{Experiments}
\label{sec:experiments}

\begin{figure*}[t]
    \centering
    \includegraphics[width=\linewidth]{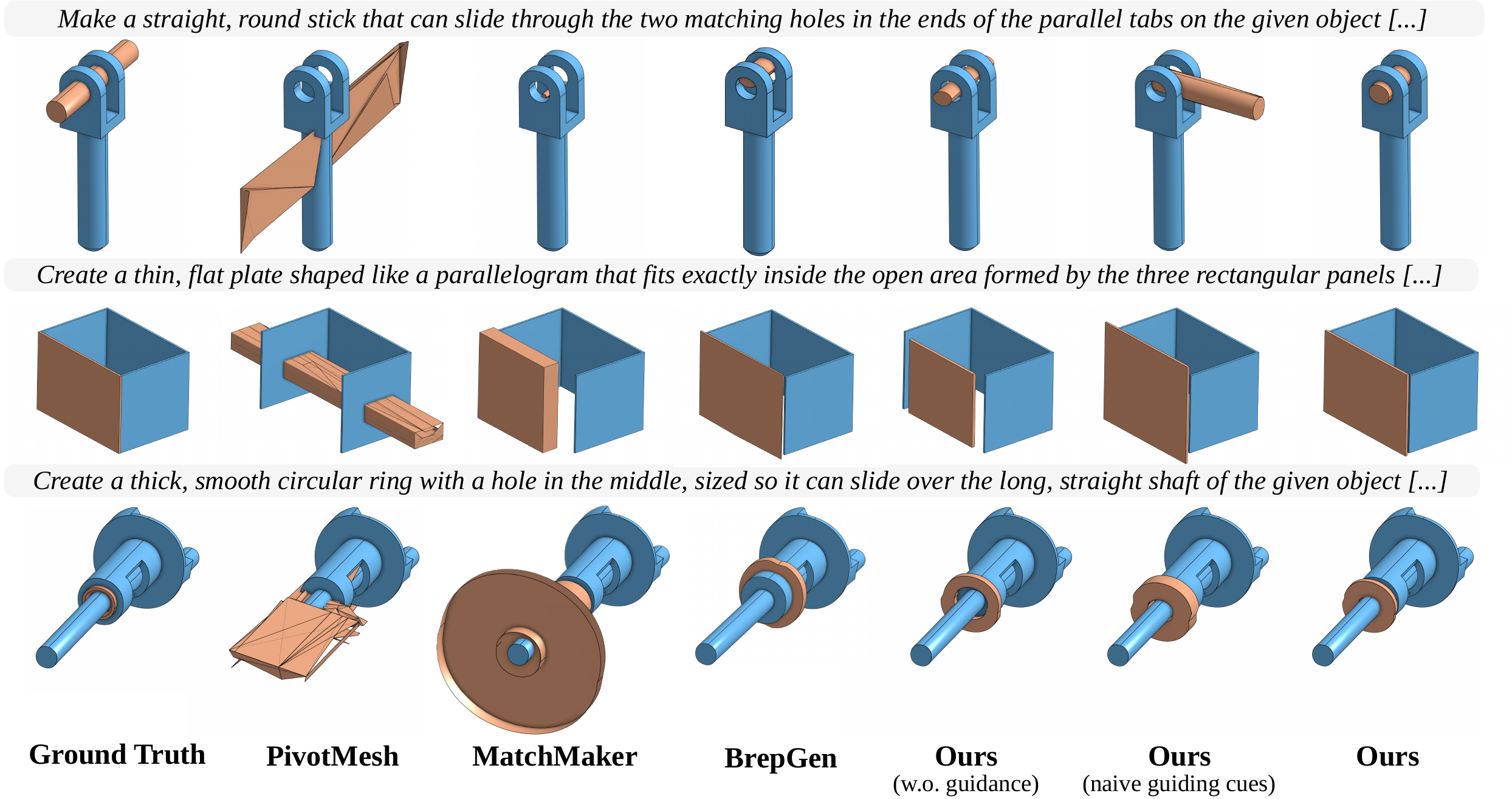}
    \caption{\textbf{Qualitative results.} We demonstrate the qualitative results of our method and the other two baselines. The condition CAD models are in \textcolor{lightblue}{\textbf{blue}} and the generated CAD models are in \textcolor{lightorange}{\textbf{orange}}.}
    \label{fig:qualitative-results}
\end{figure*}

\subsection{Implementations}
\noindent \textbf{Network \& pipeline.} Our implementation is built upon the BrepGen's pipeline~\cite{xu2024brepgen} which has four sequential stages: (I) generating face bounding boxes, (II) generating face points, (III) generating edge bounding boxes, and (IV) generating edge points. Each stage uses a diffusion model conditioned on the previous stages' output. Following these four stages, a post-processing step is performed to recover vertices and the B-rep connectivity (topology) and finally build B-rep with the OpenCASCADE~\cite{opencascade} kernel. We modify the first stage to use our proposed network and apply our proposed guidance to generate the face bounding boxes, as it carries the position and general geometric information of a B-rep model. The remaining stages of BrepGen are kept unchanged. Specifically, we represent each B-rep face as an axis-aligned bounding box $x^{(i)}\in\mathbb{R}^6$, encoded by an MLP into a $D=768$ latent feature. The noise predictor $\epsilon_\theta$ is a Transformer-based module, and the diffusion timestep is added element-wise to the noisy latent features. Text prompts are embedded with BERT~\cite{devlin2019bert} and linearly projected into the same latent space. We initialize the Transformer from the public BrepGen weights pretrained on ABC~\cite{koch2018abc}. During the sampling process, we apply guidance at $4$ late steps of the reverse diffusion process ($105$, $85$, $65$, $45$). At each guidance step, we pass the current intermediate face bounding boxes $\mathbf{x}_t$ through BrepGen's pretrained face-point diffusion model (its second stage) to obtain a $32\times32$ grid of points per face. These points need not be decoded from fully denoised bounding boxes; although such intermediate faces cannot yet be assembled into a valid B-rep, they already carry enough geometric information for our Guiding-Cue Predictor to compute the guiding cue and for the scoring function to steer the next denoising step. Once the bounding-box stage finishes, the remaining BrepGen stages run unchanged.

\noindent \textbf{Guiding-Cue Predictor.} To obtain geometric-guiding cue, we optimize on a set of face points decoded from the intermediate bounding boxes. Specifically, we first match the generated and conditional contact faces via Hungarian matching on point-to-mesh distance, then match their boundary edges via Hungarian matching on Chamfer distance. For each matched face pair, we optimize a global translation and UV-grid displacements with Adam on the total cost (Eq.~\ref{eq:shape_cost}).

\noindent \textbf{Coordinate frame \& normalization.} Each KnitCAD sample is a pair of B-rep models in their assembled state that share a single global coordinate frame, and both our method and all baselines generate the target B-rep in this shared frame. Before training, we compute one translation and scale so that the condition model is centered at the origin and lies within $[-3, 3]$ per axis, and apply the same transform to both the condition and target models.

\noindent \textbf{Training.} We train the denoiser on a single NVIDIA A100 80GB GPU using AdamW~\cite{loshchilov2017decoupled} with a learning rate of $5\times10^{-4}$, a batch size of $256$, and $5{,}000$ epochs, setting the number of generated contact faces to $M=10$. Full implementation details are provided in the Supplemental Material.

\subsection{Baselines \& Experimental Setups}

\textbf{Baselines.} We evaluate our method against MatchMaker \cite{wang2025matchmaker}, the most closely related work to ours, and BrepGen \cite{xu2024brepgen}, which we adapt from a single CAD generation baseline. We extend MatchMaker with text conditioning. For BrepGen, we integrate text and CAD conditions following their conditional generation setup.
We compare our method with a recent direct mesh generation method, namely PivotMesh~\cite{weng2024pivotmesh}. For PivotMesh, we adapt it to incorporate additional conditions. As noted in prior works~\cite{li2025caddreamer, chen2025cadcrafter}, methods trained with implicit representations of 3D shapes (e.g., point clouds, 3D mesh) often produce outputs that are unsuitable for directly reconstructing CAD models. Furthermore, we also compare our method with and without guidance.
We detail the implementation of all the baselines in the Supplemental Material. 

\noindent \textbf{Evaluation Protocol \&  Metrics.} Our method and all baselines are trained and evaluated on our proposed \textit{\datasetName{}} dataset. We then evaluate our method on test samples, generating 16 B-rep models per sample and reporting averaged metrics over the successfully built generated CAD models. Following prior works~\cite{luo2025physpart, khan2024text2cad, jayaraman2022solidgen, xu2024brepgen}, we report the following evaluation metrics: (\textit{i}) Chamfer Distance (CD) is used to measure the alignment with the ground truth shape; (\textit{ii}) Intersection Volume Percentage (IV) quantifies interpenetration under the condition of CAD models; (\textit{iii}) Proximity (PR) computes the average distance between corresponding contact faces (in $\times10$ mm); and (\textit{iv}) Valid Ratio (VR) calculates the percentage of watertight B-rep generation. Within these metrics, IV and PR assess the geometric compatibility, while CD evaluates semantic alignment with the prompt. For PivotMesh, we only report CD, as it directly generates the mesh and not the CAD model. 

\subsection{Main Results}
Table~\ref{table:compositional-text2cad-generation} compares MatchMaker, BrepGen, PivotMesh, and our method with and without guidance. Our approach achieves better overall performance on defined metrics with lower CD, lower PR, and lower IV, particularly when guidance is applied. 
The guidance introduces an empirical trade-off in which VR is approximately $2\%$ lower, while IV improves by approximately $19\%$, indicating a substantially better satisfaction of geometric constraints with a modest reduction in plausibility.

\begin{table*}[t]
    \centering
    \footnotesize     
    \begin{minipage}[t]{0.32\linewidth}
        \centering
        \begin{minipage}[c][3cm][c]{\linewidth} 
            \caption{\textbf{Scoring function analysis.}\label{tab:reward-analysis}}
            \centering
            \setlength{\tabcolsep}{2.5pt} 
            \begin{tabular}{r | cccc}
                \toprule
                \textbf{Ablation} & $\textbf{CD} \downarrow$ & $\textbf{PR} \downarrow$ & $\textbf{IV} \downarrow$ & $\textbf{VR} \uparrow$ \\ 
                \midrule
                No Guidance & 88.69 & 0.24 & 9.42 & \textbf{0.45} \\
                Scoring Func. & \textbf{83.38} & \textbf{0.23} & \textbf{7.59} & 0.44 \\ \midrule
                Scoring Func. (w.o. $D_\text{sem}$) & 87.50 & 0.26 & 8.52 & 0.43 \\
                \bottomrule
            \end{tabular}
        \end{minipage}
    \end{minipage}%
    \hfill
    \begin{minipage}[t]{0.32\linewidth}
        \centering
        \begin{minipage}[c][3cm][c]{\linewidth}
            \caption{\textbf{Predictor analysis.}\label{tab:predictor-analysis}}
            \centering
            \setlength{\tabcolsep}{2.5pt} 
            \begin{tabular}{r | cccc}
                \toprule
                \textbf{Ablation} & $\textbf{CD} \downarrow$ & $\textbf{PR} \downarrow$ & $\textbf{IV} \downarrow$ & $\textbf{VR} \uparrow$ \\ 
                \midrule
                Predictor & \textbf{83.38} & \textbf{0.23} & \textbf{7.59} & \textbf{0.44} \\ \midrule
                Predictor (w.o. $\mathcal{C}_{\text{shape}}$) & 88.81 & 0.25 & 8.72 & \textbf{0.44} \\
                Predictor (w.o. $\mathcal{C}_{\text{pos}}$) & 91.89 & 0.24 & 8.76 & 0.43 \\
                \bottomrule
            \end{tabular}
        \end{minipage}
    \end{minipage}%
    \hfill
    \begin{minipage}[t]{0.32\linewidth}
        \centering
        \begin{minipage}[c][3cm][c]{\linewidth}
            \caption{\textbf{Guiding cues comparison.} \label{table:guidance-comparison}}
            \centering
            \setlength{\tabcolsep}{2pt}
            \begin{tabular}{r | cccc}
                \toprule
                \textbf{Method} & $\textbf{CD} \downarrow$ & $\textbf{PR} \downarrow$ & $\textbf{IV} \downarrow$ & $\textbf{VR} \uparrow$ \\ 
                \midrule
                SVDD + Naive & 87.21 & 0.25 & 7.16 & 0.44 \\
                SVDD + Ours & \textbf{86.24} & \textbf{0.23} & \textbf{6.91} & \textbf{0.45} \\ 
                \midrule
                ZO + Naive & 87.12 & \textbf{0.23} & 7.65 & 0.43 \\
                ZO + Ours & \textbf{83.38} & \textbf{0.23} & \textbf{7.59} & \textbf{0.44} \\ 
                \bottomrule
            \end{tabular}
        \end{minipage}
    \end{minipage}
\end{table*}

\label{sec:generalization-analysis}
\begin{figure}[h]
    \centering
    \includegraphics[width=\linewidth]{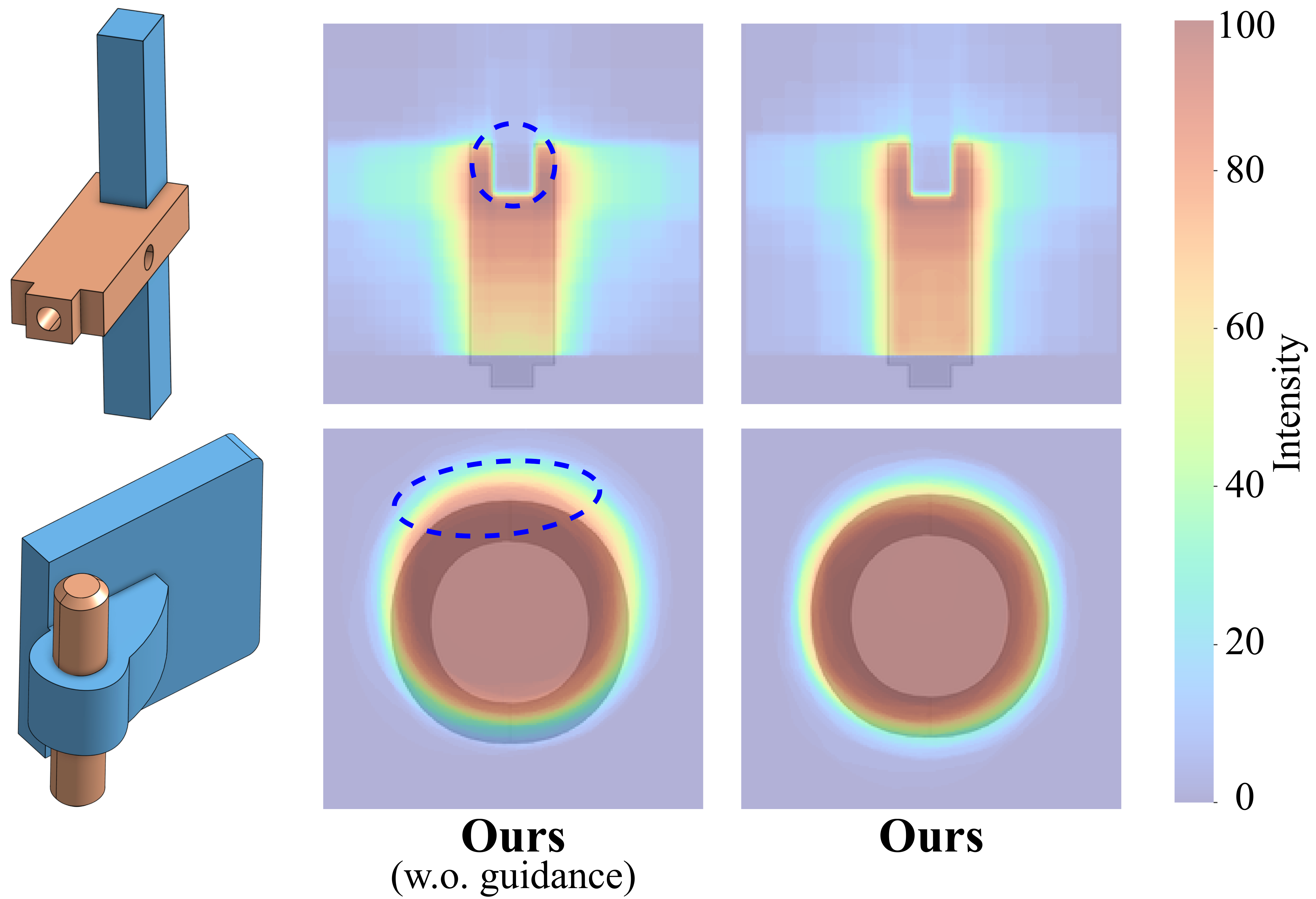}
    \caption{\textbf{Heatmap of generation.}}
    \label{fig:guidance-analysis}
\end{figure}

\subsection{Human Evaluations \& Qualitative Results}
We perform user preference studies with 20 users (aged 18–27, including designers, engineers, mechanics, and students) and 1,000 comparisons to PivotMesh, BrepGen, and MatchMaker. Participants are tasked with annotating which of two generated CAD models (or tie) is (1) more semantically aligned with the input text prompt and (2) more geometrically fit with the conditional CAD model. Table~\ref{table:user-preference-study} shows the rate of outputs from our method compared to baselines for questions (1) and (2) under ``Semantics'' and ``Geo. Compatibility'', respectively. The result indicates that MatchMaker gains a competitive semantics alignment with the highest tie rate. The remaining results show that users favor outputs from our method over those from the baselines, as evidenced by a win-rate that is $1.3$-$9.5$ times higher than the tie rate.

\begin{figure*}[t]
    \centering
    \includegraphics[width=0.97\linewidth]{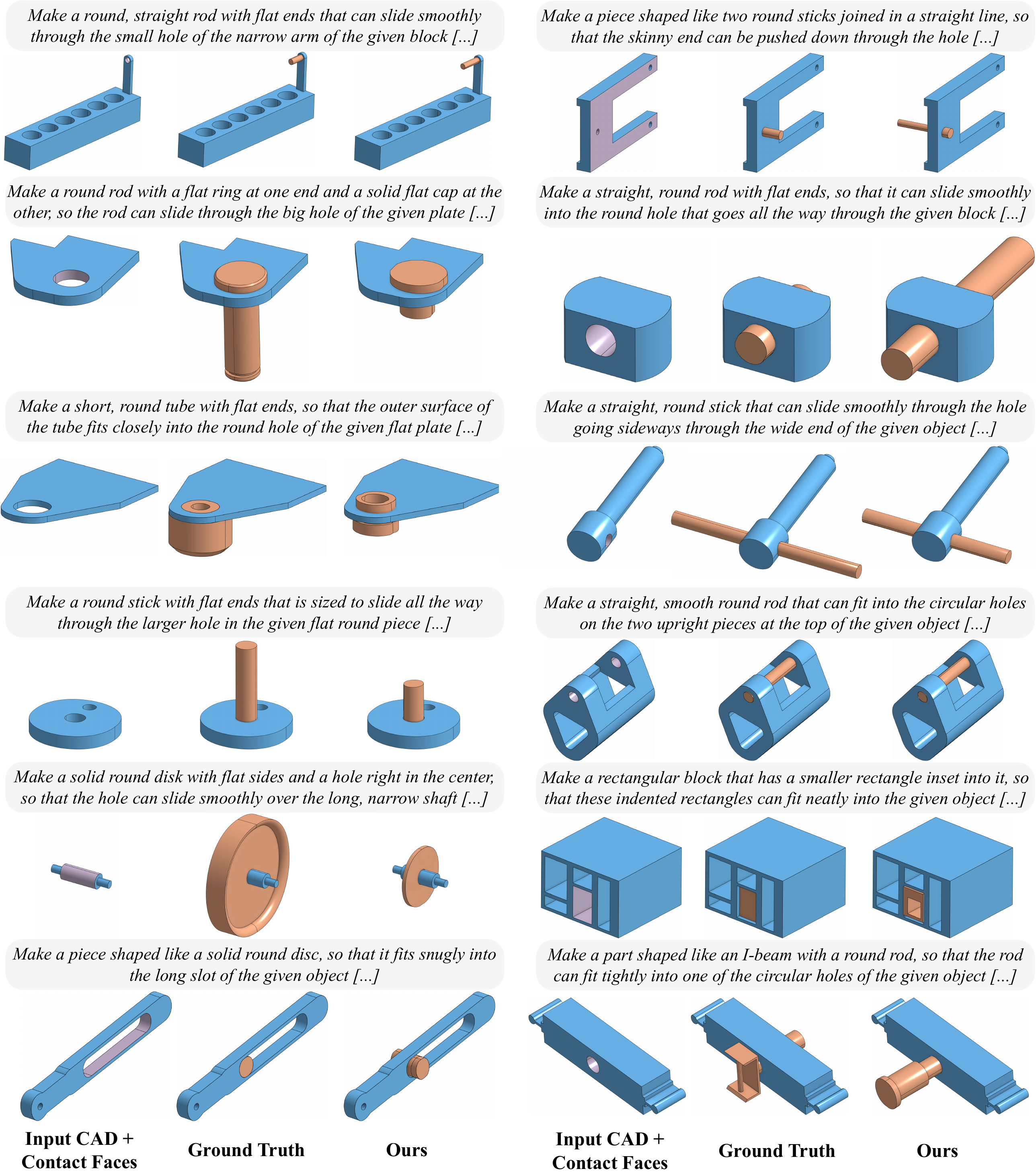}
    \caption{\textbf{Additional qualitative results.} The condition CAD models are in \textcolor{lightblue}{\textbf{blue}}, the target CAD models are in \textcolor{lightorange}{\textbf{orange}}, and the desired contact faces are in \textcolor{lightpurple}{\textbf{purple}}.}
    \label{fig:additional-qualitative-results}
\end{figure*}

Fig.~\ref{fig:qualitative-results} presents qualitative comparisons across the evaluated methods. 
While PivotMesh and MatchMaker can produce accurate output for conditional CAD models with few faces, they struggle with those having either more faces or complex contact face constraints. 
Without guidance, our method tends to produce CAD models that are undersized relative to the ground truth. In contrast, applying our guidance strategy yields models that align more accurately with the conditional CAD geometry, demonstrating improved fit and geometric fidelity. We show additional qualitative results from our method in Fig.~\ref{fig:additional-qualitative-results}.

\subsection{Generalization Analysis}
\textbf{How does our propose methods generalize to unseen data?} To evaluate the generalization of our proposed network and geometric-guiding cues, we perform a cross-set evaluation. Specifically, we train all models solely on the AutoMate subset and test them on the Fusion 360 Joint subset in KnitCAD. These two subsets are originate from two distinctly different CAD sources. Table~\ref{table:cross-set-eval} shows that our method outperforms the baselines on this unseen data and the guidance improves the overall performance via lower CD and lower IV. The analysis in Fig.~\ref{fig:guidance-analysis} further highlights that our scoring function helps generate CAD models that align more accurately. Specifically, it visualizes the top-down view of $100$ generated CAD models using our method with and without guidance.

\subsection{Scoring Function \& Guiding Cues Analysis}
\label{sec:geometric-guidance-analysis}
\textbf{How effectively do the proposed geometric-guiding cues support guidance mechanisms?} A straightforward guiding cue for our problem is the use of conditional contact faces. We compare the performance of our guiding cues against this naive cue (Naive). For generalization, we integrate them into two gradient-free guidance mechanisms: Zero-Order Search (ZO)~\cite{ma2025inference} and SVDD~\cite{li2024derivative}. Table~\ref{table:guidance-comparison} shows that our guiding cues steer the generation more effectively across all metrics and guidance mechanisms. Specifically, while the naive approach achieves competitive results in geometric compatibility, it compromises semantic alignment, as indicated by the large gap in CD. 

\begin{figure}[h]
    \centering
    \includegraphics[width=\linewidth]{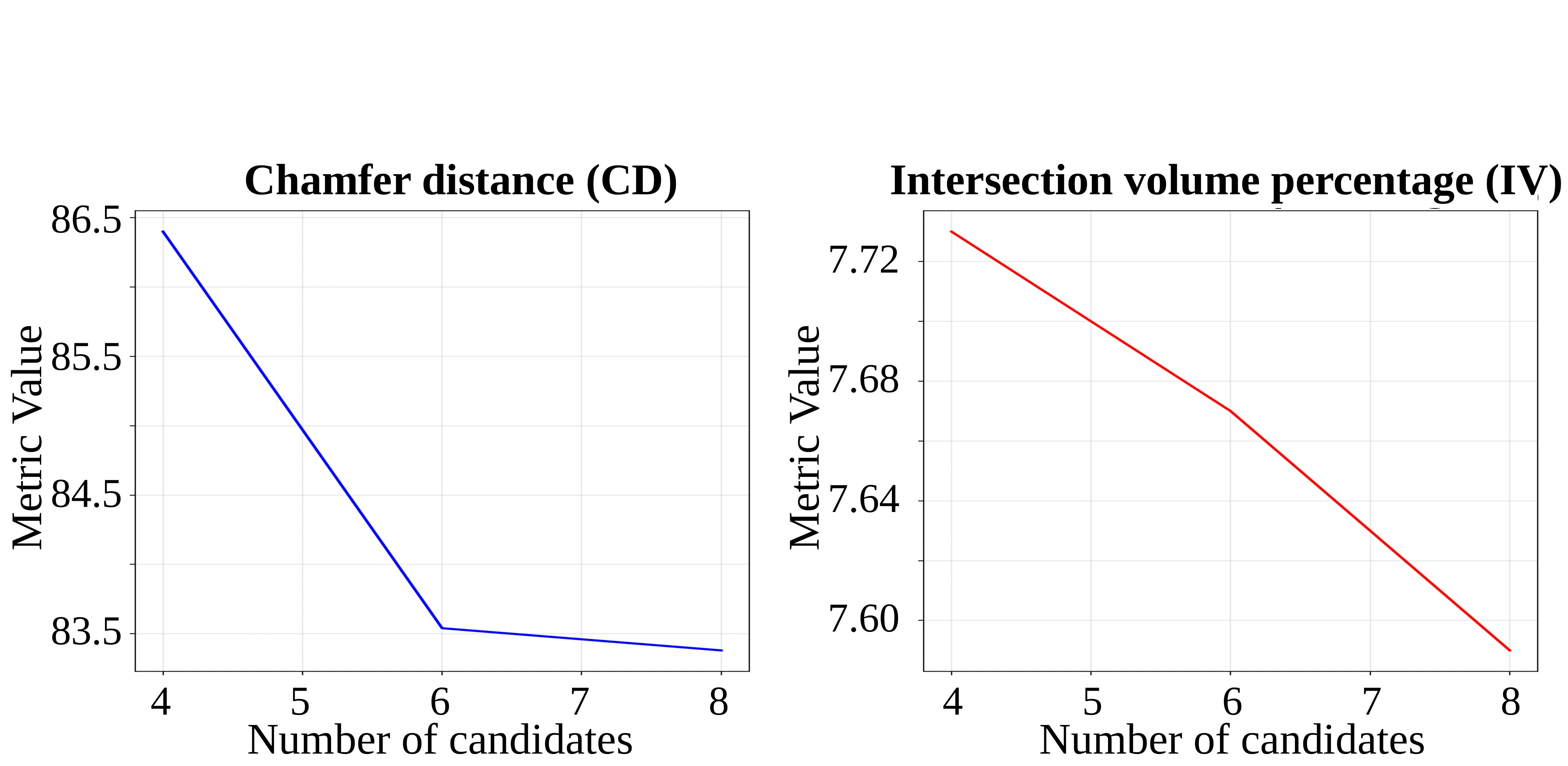}
    \caption{\textbf{Scalability preservation.} \label{fig:scalability-analysis}}
\end{figure}

\noindent \textbf{Does our proposed scoring function preserve the scalability of the guidance mechanism?} Since Zero-Order Search~\cite{flaxman2004online} scales with the number of candidates $N_p$, we analyze the performance of our scoring function across different values of $N_p$. The analysis in Fig.~\ref{fig:scalability-analysis} confirms that our approach successfully preserves this scalability. Specifically, increasing the number of candidates consistently improves metrics such as CD and IV.

\subsection{Ablation Studies}
\label{sec:ablation-studies}
\noindent \textbf{Semantic Preservation Term Analysis.} Table~\ref{tab:reward-analysis} demonstrates that the semantic preservation term plays a crucial role in guiding the generation. While relying solely on the geometric term (w.o. $D_\text{sem}$) improves CD and IV compared to the baseline, it can disrupt the overall semantic structure, leading to a decrease in VR. Integrating $D_\text{sem}$ into the full scoring function mitigates this issue and significantly improves the CD score.

\noindent \textbf{Guiding-Cue Predictor Analysis.} Table~\ref{tab:predictor-analysis} shows that all cost functions in the optimization play an important role in predicting the geometric guiding cues. Specifically, removing either $\mathcal{C}_{\text{shape}}$ or $\mathcal{C}_{\text{pos}}$ affects semantic fidelity, as the cues lose their geometric-guiding information, leading to degraded CD.

\section{Discussions}
\subsection{Broader Impact}
\begin{figure}[ht]
    \centering
    \includegraphics[width=\linewidth]{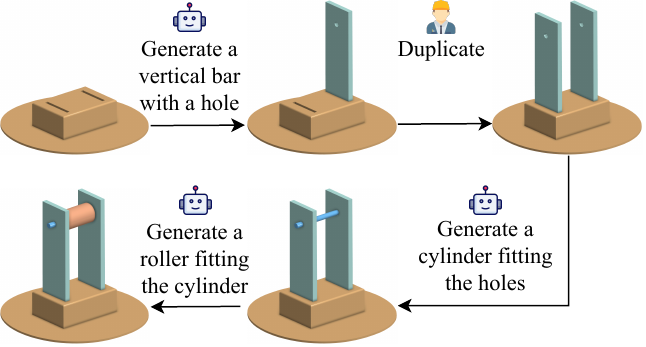}
    \caption{\textbf{Sequential CAD generation.}}
    \label{fig:multi-step}
\end{figure}

We believe that our work can serve as a step towards real-world applications. One such application is the automated generation of multiple parts in a design, as shown in Fig.~\ref{fig:multi-step}. As a preliminary demonstration, we integrate our model in a design process where new parts are sequentially generated given a single base, in-the-wild CAD model. Our method can also be integrated as a plug-in into existing 3D CAD software to support the designer during their daily work.

\subsection{Failure cases}
\begin{figure}[ht]
    \centering
    \includegraphics[width=\linewidth]{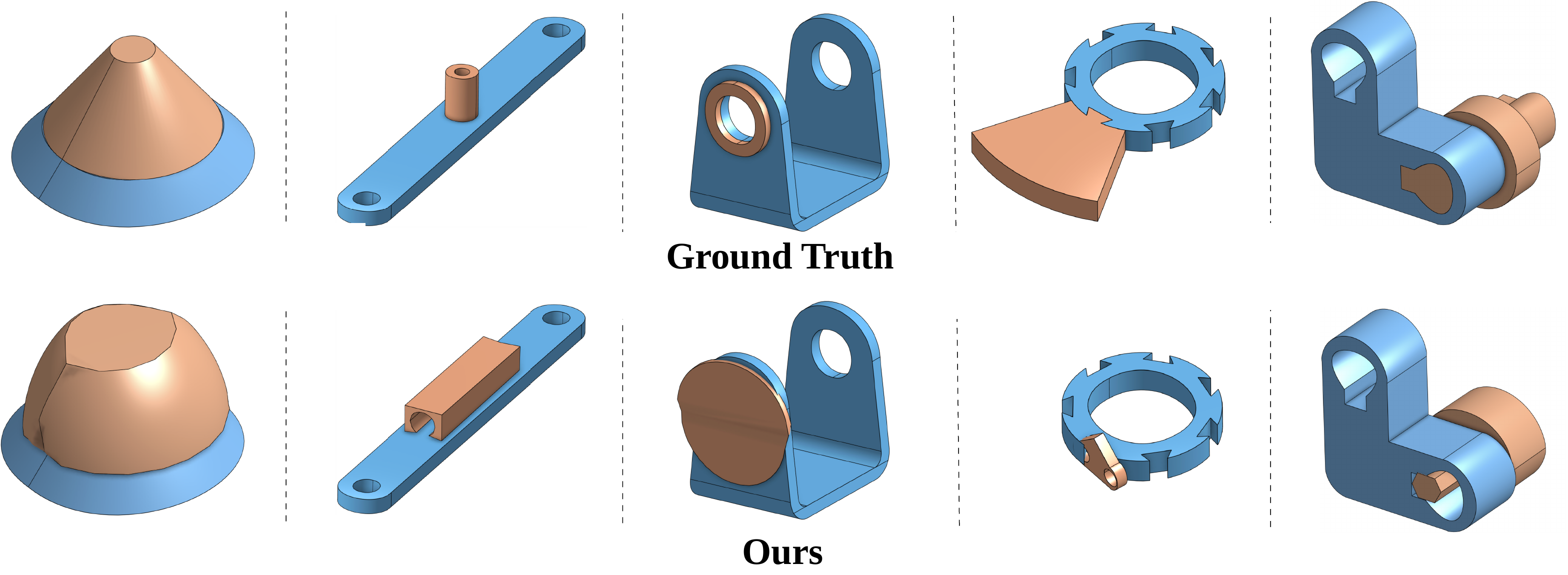}
    \caption{\textbf{Failure cases.}}
    \label{fig:failure-cases}
\end{figure}

Besides the failure modes inherited from the BrepGen backbone, we observe several additional failure cases, as shown in Fig.~\ref{fig:failure-cases}.
First, contact faces do not always provide sufficient geometric constraints for conditioned CAD generation. For example, in the third sample of Fig.~\ref{fig:failure-cases}, the inner diameter of a generated ring should match the diameter of the corresponding hole in the conditioning CAD model, yet this geometric relationship is not encoded by the contact faces alone.
Second, generation quality degrades when the conditioning contact faces are complex, such as mixed or freeform surfaces. We attribute this limitation to the underrepresentation of these contact face types in the training dataset, resulting in insufficient examples for the model to learn robust semantic priors.

\subsection{Limitations}
Despite promising results, our method still has limitations.
First, our framework generates parts sequentially, conditioning each new part on only a single previously generated part. Extending it to support multi-part conditioning would broaden its applicability to more general CAD design scenarios.
Second, while our geometric-guiding cues improve the overall performance, they perform worse in cases where the assumption of boundary-edge length equality does not hold. As detailed in the Supplemental Material, although our approach still outperforms existing baselines in these specific scenarios, our unguided variant actually yields better results than the guided one. We identify these challenges as important directions for future work to further enhance compositional CAD generation.

\subsection{Conclusions}
In this work, we introduce a new task, compositional CAD generation. To address this task, we present \textit{\datasetName{}}, a new dataset with metadata and a textual prompt for facilitating research in this direction. We propose a new network and a scoring function to enforce generated CAD models to align with the textual prompt and can be assembled with existing CAD models. Empirically, the proposed method outperforms other baselines with a clear margin, and our scoring function further preserves the scalability of the derivative-free guidance mechanism. Our code and dataset will be released to encourage future research on generating CAD models for real-world applications.

\balance
\bibliographystyle{ACM-Reference-Format}
\bibliography{main}

\clearpage
\appendix

This Supplementary Material provides extra material for the paper ``\textit{CADKnitter: Compositional CAD Generation from Text and Geometry Guidance}". The material is organized as follows:
\begin{itemize}
    \item Section~\ref{sec:sup-dataset-stats} provides details of the automatic dataset annotation pipeline and statistics.
    \item Section~\ref{sec:statistical-analysis} provides additional details of the statistical analysis on boundary edge length.
    \item Section~\ref{sec:sup-implement-ours} provides the detailed implementation of our method.
    \item Section~\ref{sec:sup-implement-baselines} provides the detailed experiment setups, including the implementation of evaluation metrics, baselines, and derivative-free guidance mechanisms.
    \item Section~\ref{sec:sup-additional-qualitative} presents more results, including additional guidance analysis, robustness analysis, and more qualitative results.
    \item Section~\ref{sec:sup-practical-application} provides details of the demonstration of practical application.
\end{itemize}

\section{Additional Details on KnitCAD Dataset}
\label{sec:sup-dataset-stats}
\subsection{Annotation Pipeline}

\begin{figure}[h]
    \centering
    \includegraphics[width=\linewidth]{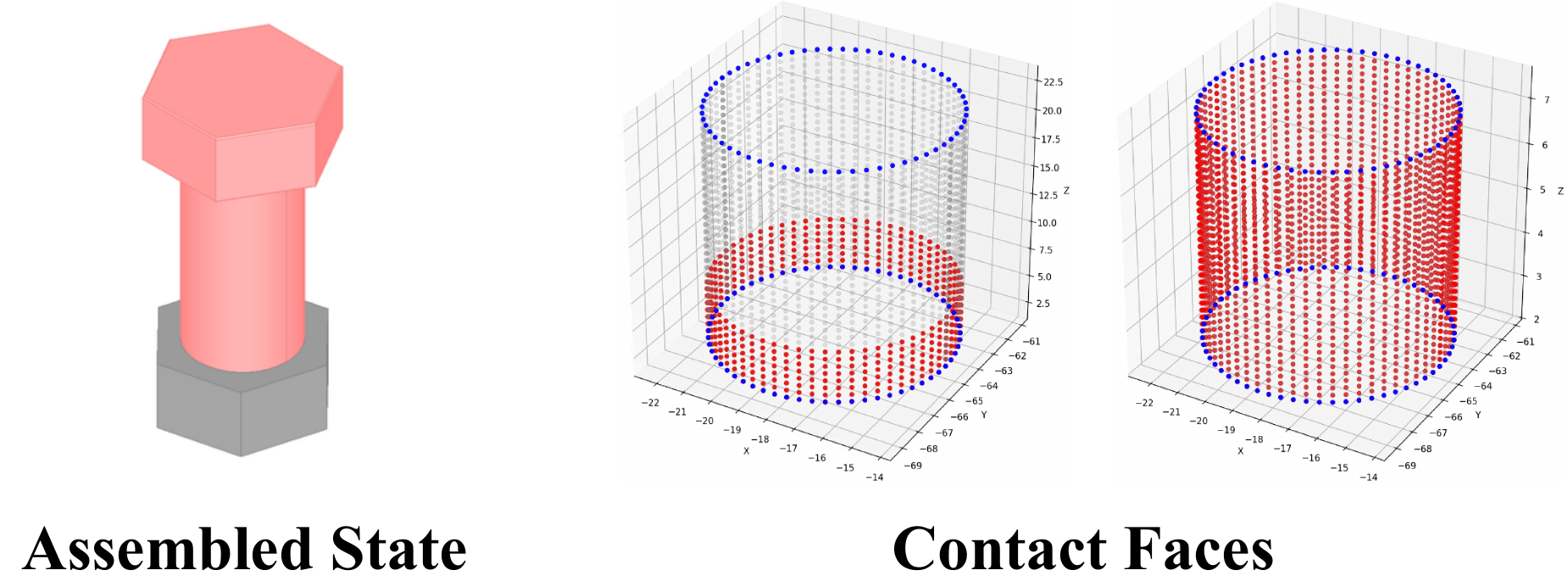}
    \caption{\textbf{Contact faces}. Left: Two CAD models in their assembled state. Right: Point sets extracted from the corresponding contact faces for the \textcolor{pink}{pink} and \textcolor{gray}{gray} parts. Points satisfying the contact conditions and belonging to boundary edges are highlighted in \textcolor{red}{red} and \textcolor{blue}{blue}, respectively.}
    \label{fig:sup-contact-face}
\end{figure}

\noindent\textbf{Contact Face Labeling}. For each pair of B-rep models, we first assemble the two parts using the joint metadata from Fusion 360 Joint~\cite{willis2021joinable} and Automate~\cite{jones2021automate}. We set the distance threshold to $\delta = 0.1$, so a point is considered close to a face if it lies within $0.1\,\mathrm{mm}$ of that face. Fig.~\ref{fig:sup-contact-face} shows an example of the contact faces and the points from the two objects that satisfy the contact conditions defined in Sec. 3.2 in the main paper. 

\noindent \textbf{Shape description and textual prompt generation}. We resize all multi-view images to a resolution of $512 \times 512$ before providing them, along with the prompts, to the MLLM. We show the prompts that we use with the MLLM to describe the shape of each CAD model and generate the compositional text prompts for each pair of CAD models in Fig.~\ref{fig:sup-prompt-shape-description} and Fig.~\ref{fig:sup-prompt-compositional}.

\begin{figure*}[h]
\centering
\begin{tcolorbox}[title=Prompt synthesizing shape description]
\small

You are given four images in a grid of 2x2 of a single object. Your task is to write a description for the object from these images.

\medskip

Follow these rules strictly:
\begin{itemize}
  \item Goal: Produce a precise, compact description of geometry and topology (not color, brand, text, background, lighting, or materials).
  \item Views: Use all provided views; reconcile conflicts. Do not describe the object from a single specific reference view (top, left, right, bottom).
\end{itemize}

\medskip

Output: A detailed description of the object's geometry and topology. Do not mention features that are not visible in all four images.

\medskip

Examples:
\begin{itemize}
  \item ``The object is a rectangular cuboid with six faces. Two opposite faces are open rectangular frames, each with rounded-rectangle cutouts along all four edges. The remaining four faces are solid panels. Each solid panel has a row of evenly spaced, elongated rounded-rectangle holes parallel to the long axis of the cuboid.''
  \item ``The object is a thin, square plate with rounded edges and corners. One face of the plate is solid and featureless. The opposite face features a recessed square grid, forming a pattern of evenly spaced square openings bounded by a raised border along all four edges. The thickness of the object remains consistent throughout, and the grid pattern occupies the entire face except for the surrounding raised border.''
\end{itemize}

\end{tcolorbox}
\vspace{-8pt}
\caption{Prompt used to synthesize shape description from multi-view images.}
\label{fig:sup-prompt-shape-description}
\vspace{-12pt}
\end{figure*}

\begin{figure*}[h]
\centering
\begin{tcolorbox}[title=Prompt synthesizing instructions]
\small

The image shows two CAD models that fit together. The engineer is only given the condition model (not the image or descriptions). Your task is to act as a layperson and write a short prompt (1--2 sentences) for the engineer to design the target model so that it can connect with the condition model.

\medskip

The prompt should clearly describe the shape and structure of the target object, and specify which part of it will attach or fit into the condition object. Do not mention colors, functions, or purposes of the object. Use simple, everyday words and avoid vague or technical terms.

\medskip

Example prompts:
\begin{itemize}
  \item ``Make a piece shaped like a flat circle with a hole in the middle, so it can slide snugly onto the tall round stick of the given object and rest flush against the wide round part at the base.''
  \item ``Generate a long, straight stick with six flat sides that smoothly bends into a nearly full open hook at one end, so that the curved hook can wrap snugly around the round cylinder of the given object and the straight part sits flat against its side.''
\end{itemize}

\medskip

Use the following descriptions to identify the objects in the image and generate the prompt:

\noindent
Condition Object Description: \{\texttt{condition object description}\}

\noindent
Target Object Description: \{\texttt{target object description}\}

Prompt for the engineer:

\end{tcolorbox}
\vspace{-8pt}
\caption{Prompt used to synthesize instructions for generating complementary CAD parts.}
\label{fig:sup-prompt-compositional}
\end{figure*}

\subsection{Dataset Statistics}
\begin{figure*}[h]
    \centering
    \begin{subfigure}{0.47\linewidth}
        \centering
        \includegraphics[width=\linewidth]{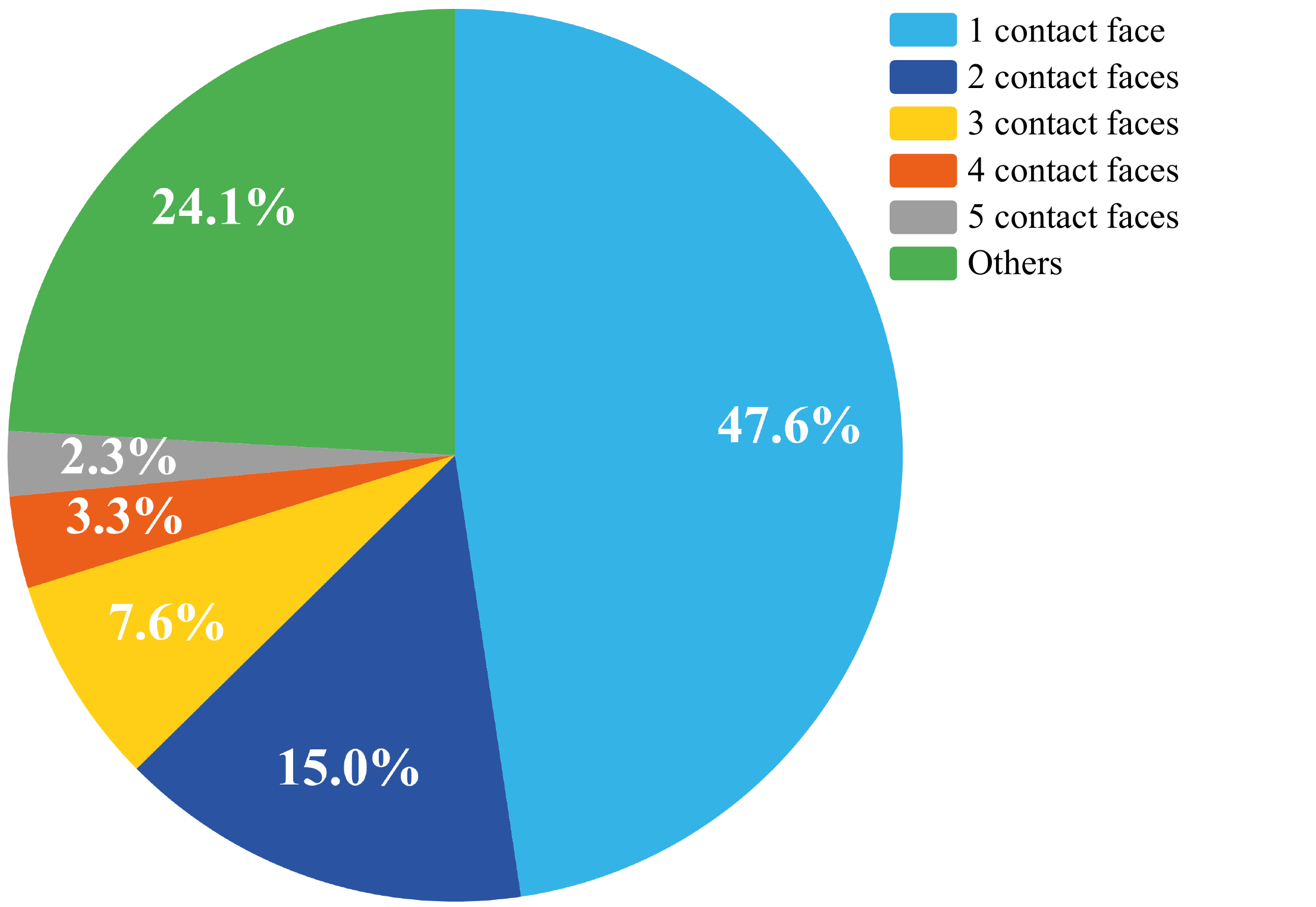}
        \caption{Distribution of condition types.}
        \label{fig:sup-dataset-stats:contact}
    \end{subfigure}
    \hfill
    \begin{subfigure}{0.52\linewidth}
        \centering
        \includegraphics[width=\linewidth]{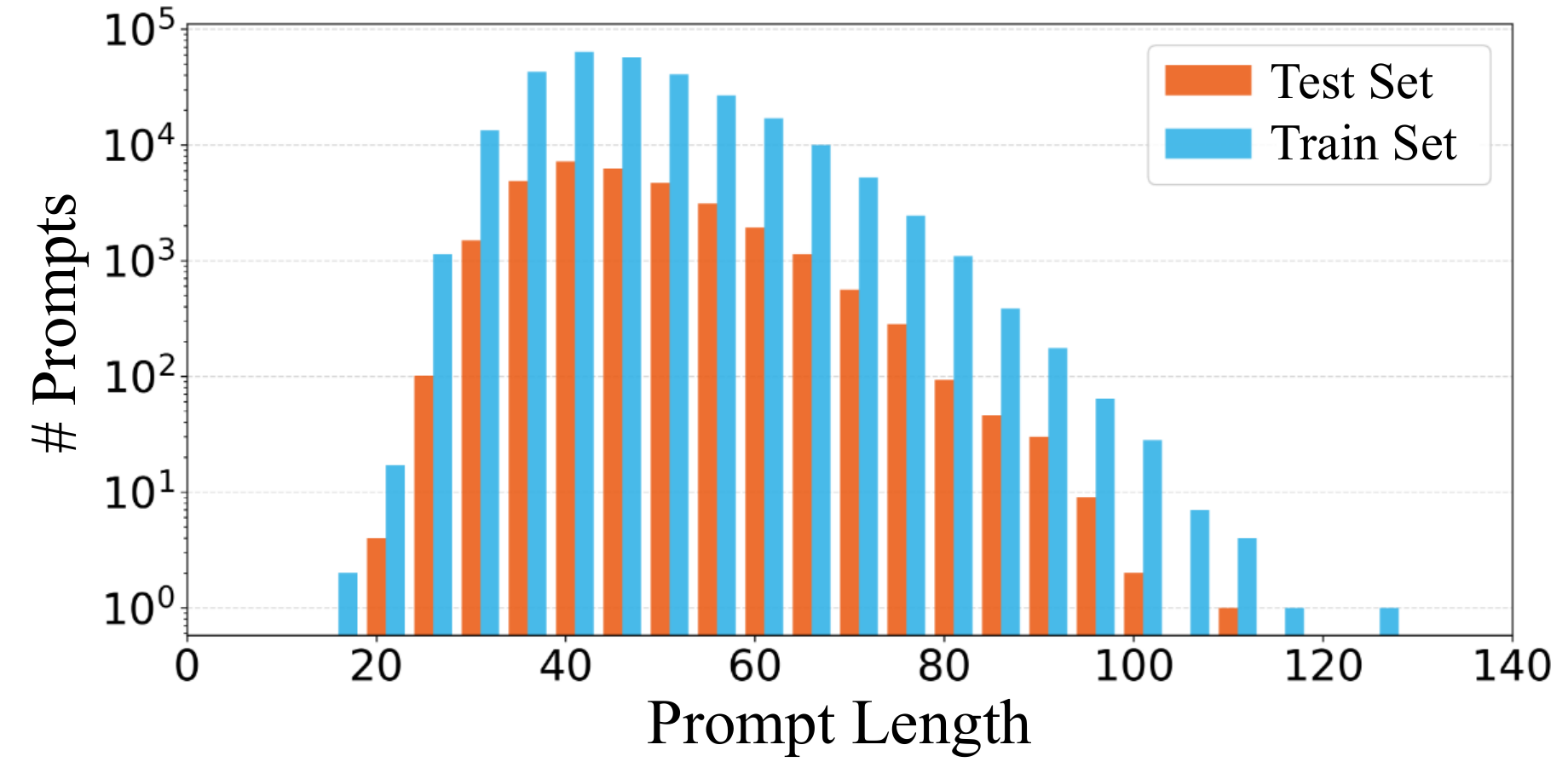}
        \vspace{0.5pt}
        \caption{Distribution of text prompt lengths.}
        \label{fig:sup-dataset-stats:text}
    \end{subfigure}
    \vspace{-4pt}
    \caption{\textbf{Dataset statistics.} (a) Distribution of contact-face condition types. (b) Distribution of text prompt lengths.}
    \label{fig:sup-dataset-stats}
\end{figure*}

\textbf{Condition Types}. We categorize the dataset samples by the number of conditional contact faces, with the full distribution shown in Fig.~\ref{fig:sup-dataset-stats:contact}. On average, each conditional CAD model contains 1.55 (1-2) contact faces. 

\noindent\textbf{Prompt Length Distribution}. Fig.~\ref{fig:sup-dataset-stats:text} illustrates the distribution of text prompt lengths across the dataset. The prompts range from 13 to 160 words, with an average length of 47.81 words.

\vspace{-8pt}
\subsection{Dataset Details}
\textbf{Train–test Split}.  
For Fusion 360 Joint, we use the official train–test split provided with the dataset. For Automate, we partition the dataset into training, validation, and test sets using a 60/20/20 ratio. To prevent data leakage, this split is performed at the pair level; thus, all samples derived from the same pair of objects are restricted to the same subset.

\noindent\textbf{Data Representation}.  
For each pair of B-rep models, we first transform both models into their assembled state. We then sample UV faces and edges on each B-rep model in this assembled configuration. Each data sample in KnitCAD consists of a pair of B-rep models that are already assembled and share the same global coordinate frame. Our method and all baselines are trained to generate B-rep models in this shared global coordinate frame.

\section{Details of Boundary Edge Length Statistical Analysis}
\label{sec:statistical-analysis}
We randomly select 32,000 samples from the Automate subset in KnitCAD. For each sample, we extract the boundary edges on contact faces from two CAD models. To compare these boundary edges, we calculate the relative difference for each pair using the formula:
\begin{equation}
    \text{Relative Difference} = \frac{|L_a - L_b|}{\max(L_a, L_b)}\quad,
\end{equation}
where $L_a$ and $L_b$ are the lengths of the edges from the first and second CAD models, respectively. Finally, we report the distribution and correlation of the boundary edge pairs that exhibit the smallest relative difference. We observe that small relative differences (e.g., approximately 5\%) are primarily attributed to samples where the contact faces do not perfectly overlap, but remain within a 0.1 mm distance threshold.

\section{CADKnitter Implementation Details}
\label{sec:sup-implement-ours}
\subsection{Compositional CAD generation network}
\noindent\textbf{Implementation Details}. 
In our work, the set of face entities is denoted by $\mathbf{x} = \{x^{(i)}\}_{i=1}^{N}$, where each $x^{(i)} \in \mathbb{R}^6$ is an axis-aligned bounding box. Each bounding box $x^{(i)}$ encloses one face and is represented as
\[
x^{(i)} = [x_{\text{min}}, y_{\text{min}}, z_{\text{min}}, x_{\text{max}}, y_{\text{max}}, z_{\text{max}}],
\]
which stores the coordinates of the bottom-left and top-right corners.
We encode each bounding box $x^{(i)}$ using an MLP that maps the 6D coordinates to a latent feature with hidden dimension $D$. The diffusion timestep $t$ is encoded and added element-wise to the noisy latent feature. The noise predictor $\epsilon_\theta$ is implemented as a Transformer-based module. After denoising, another MLP maps the latent features back to bounding box coordinates.

For the text input, we extract embeddings using a text encoder, specifically BERT~\cite{devlin2019bert}. We then apply a linear projection to map the text embeddings into the same latent space as the B-rep face bounding box embeddings. All latent features of both face bounding boxes and texts use the same hidden dimension $D = 768$.
 
To extract the face point sets, we leverage the second stage in BrepGen. Given the bounding boxes $\mathbf{x}$, we use the pretrained face point diffusion model to get the grid face points
\[
s = \{\mathbf{p}_{1}, \mathbf{p}_{2}, \dots, \mathbf{p}_{N_{s}}\} \in \mathbb{R}^3,
\]
where $N_{s}=32\times32$ is the number of sampled points on the face.

\textbf{Training}. Following~\cite{xu2024brepgen, jayaraman2022solidgen}, we split closed faces (e.g., cylindrical faces) along their seams (e.g., a closed cylinder is split into two four-sided half-cylinders). To keep training within GPU memory limits, we filter out B-rep models that have more than 70 faces, more than 10 contact faces, more than 40 edges per face, or that are composed of multiple solid bodies. After this filtering, we obtain 95{,}004 samples to train the face bounding-box denoiser.
Before training, we normalize B-rep models in the shared global coordinate frame. For each pair of B-rep models, we compute a translation and a scaling factor so that the condition B-rep model is centered at the origin and lies within the range $[-3, 3]$ along each axis. We then apply this same translation and scaling to both the condition and the target B-rep models. For training,
we set the number of contact faces for generated CAD models to $M = 10$. We train the denoiser on a single NVIDIA A100 80GB GPU and use half-precision to reduce memory usage and speed up training. We use AdamW~\cite{loshchilov2017decoupled} with a learning rate of $5\times 10^{-4}$, a batch size of 256, and train the diffusion model for 5{,}000 epochs. We initialize our model using the public BrepGen weights pre-trained on the ABC dataset.

\vspace{-4pt}
\subsection{Geometric Enforcement}
Following~\cite{yuan2023physdiff, wang2023diffusebot}, we apply guidance at a few late steps of the reverse diffusion process. We set the weight for semantic preservation term in computing candidate score to $\lambda_\text{sem} = 10^4$. For the FGW in Eqn.3, we set $\lambda=0.5$ as in~\cite{vayer2018optimal}. We evaluate different numbers of guidance steps, applied from $t = 105$ down to $t = 45$. As shown in Table~\ref{table:compositional-text2cad-generation}, using 4 guidance steps gives the best overall performance.

\begin{table}[h]
    \centering
    \renewcommand
    \tabcolsep{6.5pt}
    \resizebox{0.9\linewidth}{!}{
        \begin{tabular}{c | cccc}
            \toprule
            \textbf{No. Guidance Steps} & $\textbf{CD} \downarrow$ & $\textbf{PR} \downarrow$ & $\textbf{IV} \downarrow$ & $\textbf{VR} \uparrow$ \\ 
            \midrule
            2 & 88.72 & 0.26 & \textbf{6.53} & \textbf{0.44} \\
            4 & \textbf{83.38} & \textbf{0.23} & 7.59 & \textbf{0.44} \\
            8 & 86.58 & 0.24 & 7.43 & 0.43 \\
            \bottomrule
        \end{tabular}
    }
    \caption{\textbf{Effect of the number of guidance steps.} We compare different numbers of guidance steps during the reverse diffusion process.}
    \label{table:compositional-text2cad-generation}
    \vspace{-24pt}
\end{table}

\subsection{Guiding-Cue Predictor}
\label{sec:imp-guiding-sample-predictor}
The guiding-cue predictor returns a set of optimized contact faces for the generated CAD model. These optimized faces provide cues for derivative-free guidance mechanisms to steer the sampling process to follow the geometric and semantic constraints. 
However, defining exact contact-face constraints between two CAD models is nontrivial~\cite{jones2021automate}. To make the problem tractable, in implementation, we optimize one generated contact face to geometrically align with one conditional contact face. Under this setting, the guiding-sample predictor has two stages: (1) find matching face pairs and edge pairs between the generated and condition contact faces; (2) optimize the generated faces using the condition faces as references.

\begin{figure}[ht]
    \centering
    \includegraphics[width=\linewidth]{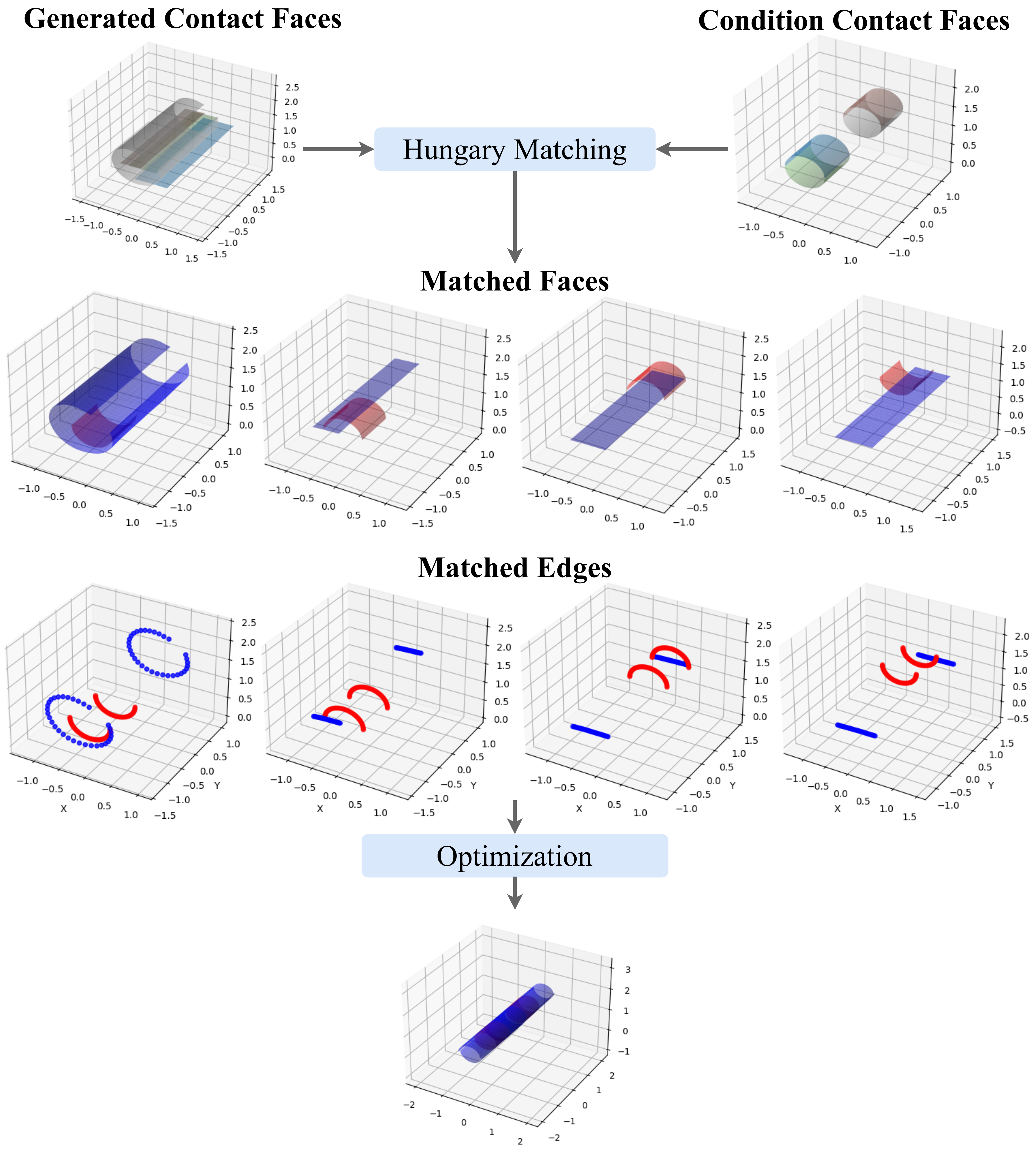}
    \caption{\textbf{Matching example.} Illustration of face and edge matching for optimization. From the generated contact faces, we use Hungarian matching to find their corresponding condition contact faces. The boundary edges of the condition contact faces are then matched to boundary edges of the generated contact faces. The \textcolor{red}{red} and \textcolor{blue}{blue} point sets represent boundary edges from the condition and generated contact faces, respectively.}
    \label{fig:matching-example}
\end{figure}

\textbf{Face and Edge Matching}.  
We first establish correspondences between the two sets of contact faces using Hungarian matching~\cite{kuhn1955hungarian}. The cost matrix is defined by the point-to-mesh distance for every pair of faces. For point-to-mesh distance, we use the implementation from PyTorch3D~\cite{ravi2020accelerating}.
For each matched face pair, we match the boundary edges of the condition contact face to the boundary edges of the generated contact face, again using Hungarian matching~\cite{kuhn1955hungarian}. In this step, the cost matrix is defined by the Chamfer Distance~\cite{ravi2020accelerating} between every pair of boundary edges. We extract boundary edges by removing edges that belong to more than one face. An illustration of the face and edge matching is shown in Fig.~\ref{fig:matching-example}. 

\textbf{Optimization}. For each matched face pair, we optimize a global translation $\mathbf{t} \in \mathbb{R}^3$, a displacement vector for each row of the face's UV grid $\mathbf{d}^u \in \mathbb{R}^{H \times 3}$, and a displacement vector for each column $\mathbf{d}^v \in \mathbb{R}^{W \times 3}$, where $H = W = 32$ is the UV-grid resolution. The position of each face point at grid location $(i, j)$ is updated as:
\begin{equation}
    \textbf{p}_{(i,j)} \leftarrow \textbf{p}_{(i,j)} + \mathbf{d}^u_i + \mathbf{d}^v_j + \mathbf{t}.
    \label{eq:sup-optimization}
\end{equation}
The key property of this parameterization is that $\mathbf{d}^u_i$ is shared by all 32 points in row $i$, including both boundary and interior points, and $\mathbf{d}^v_j$ is shared by all 32 points in column $j$. Since $\mathcal{C}_{\text{shape}}$ is computed on the boundary edges, it optimizes these shared displacements, which simultaneously moves every point in the same row or column by the same amount. This ensures that interior points move together with their corresponding boundary edge points, preserving the relative internal structure of the face.

We optimize these parameters using Adam with an alternating schedule. Every 50 optimization steps, we first update only $\mathbf{t}$ using $\mathcal{C}_{\text{pos}}$ for 5 steps to bring the contact face into proximity with the condition, then update only $\mathbf{d}^u$ and $\mathbf{d}^v$ using $\mathcal{C}_{\text{shape}}$ for the remaining 45 steps to refine the boundary shape.

We also observe that when splitting closed faces following standard B-rep generation practices~\cite{xu2024brepgen, jayaraman2022solidgen, li2025dtgbrepgen}, the faces remain in contact but their relative orientations can become misaligned. Due to this, we set $\lambda_{\text{angle}} = 0.0$ for our dataset.

\section{Details of Experimental Setups}
\label{sec:sup-implement-baselines}
\subsection{Evaluation Metrics}
In this section, we provide more details about each evaluation metric used in the paper. To ensure the reliability of our results, we run all experiments twice and report the mean.

\begin{itemize}
    \item \textbf{Chamfer Distance (CD)}. Following~\cite{xu2024brepgen}, we sample 2{,}000 points from the surfaces of both the generated and ground-truth CAD. The Chamfer Distance is computed as the average minimum distance between the two sets.
    \item \textbf{Intersection Volume Percentage (IV)}. We compute the volume of the intersection solid between the generated object and the condition object. This metric is only defined for watertight objects.
    \item \textbf{Proximity (PR)}. We first identify the faces that are in contact with the desired contact faces of the condition CAD model by applying the contact face conditions defined in the main paper. We then compute PR as the minimum distance between these contact faces on the generated and condition CAD models.
    \item \textbf{Valid Ratio (VR)}. We use OpenCASCADE~\cite{opencascade} as the CAD kernel to construct the B-rep models. A CAD model is counted as valid if OpenCASCADE can successfully build it. VR is the fraction of valid models among all generated samples.
\end{itemize}

\subsection{Implementation Details of Baselines}
In this section, we describe the implementation details of the baselines.

\textbf{Text- and CAD-conditioned version of BrepGen}~\cite{xu2024brepgen}. BrepGen is a multi-stage single CAD generation framework. It includes two Variational Auto-encoder (VAE) models. We follow the conditional generation setups in the original paper. Specifically, we train the diffusion model to generate face bounding boxes in BrepGen pipeline. For conditioning, we use the same text encoder in our method implementation to extract embeddings, while the conditional CAD faces are encoded with BrepGen's MLP encoding layer. Learnable embeddings marking desired contact faces are injected into their corresponding face representations in the same manner as our method for fair comparison. All the conditional embeddings are fused via element-wise averaging into a single conditional vector. Following the original implementation, this embedding is added to all noisy face embeddings to guide the diffusion process.

\textbf{Text-conditioned version of MatchMaker}~\cite{wang2025matchmaker}. MatchMaker is a three-stage framework: (1) contact surface extraction, (2) shape completion, and (3) clearance specification. In our problem, we already provide the contact surface labels in the dataset, and we do not use the clearance specification stage. Therefore, we only use the second stage, which performs shape completion using CAD autocompletion~\cite{xu2024brepgen}. MatchMaker uses BrepGen~\cite{xu2024brepgen} as the backbone and adapts RePaint~\cite{lugmayr2022repaint} to generate complementary geometry for the extracted contact surface. In this way, the geometric constraints are directly enforced during sampling. Similar to our method, we train the first diffusion model of BrepGen with text conditioning. The text condition is injected into the noisy latent features in the same way as in our model.
We train the text-conditioned MatchMaker for 5{,}000 epochs, with the same training setup as our method. During inference, we apply RePaint~\cite{lugmayr2022repaint} at all timesteps $t > 100$. We do not apply RePaint at smaller timesteps, since we observe that using RePaint at very small $t$ leads to less plausible generations and a lower VR compared to the results reported in the main paper.

\noindent \textbf{Text- and Mesh-conditioned version of PivotMesh}~\cite{weng2024pivotmesh}. PivotMesh is an auto-regressive model that first generates a coarse mesh representation (pivot vertices), and then refines it to a full mesh in a coarse-to-fine manner. As in our method, we first normalize the condition and target meshes using the same translation and scaling, so that the condition mesh is zero-centered and lies inside a canonical cube $[-1, 1]^3$. 
PivotMesh discretizes mesh vertices using 7-bit quantization. To support this, we further scale all meshes into a unit cube using a global scale factor computed from the maximum and minimum coordinates over the training set. In practice, we use the 90th percentile of the maximum and minimum coordinates to ignore very large outlier meshes and reduce vertex collapse for nearby vertices.
For conditioning, we introduce text tokens and condition-mesh tokens into the generation process via extra cross-attention layers in the Transformer blocks of PivotMesh. We use the pretrained auto-encoder from the original PivotMesh model, and train the autoregressive Transformer from scratch for 165{,}000 steps with a batch size of 6 on 4 NVIDIA A100 80GB GPUs. We use gradient accumulation with 2 steps, set the learning rate to $1\times 10^{-4}$, and use 2{,}000 warm-up steps.

\subsection{Implementation Details of Derivative-free Guidance Mechanisms}
\noindent \textbf{Zero-Order Search}~\cite{ma2025inference}. We implement this guidance mechanism as presented in the main paper. We choose the candidate with the lowest score $R(\textbf{S}_t^{(n)})$ as the next intermediate step.

\noindent \textbf{Soft Value-based Decoding in Diffusion Models (SVDD)}~\cite{li2024derivative}. SVDD aims to optimize a downstream reward function while preserving the naturalness of the generated samples. In our context, our scoring function serves as the reward function, and the validity of the generated B-rep samples represents this naturalness. We follow Algorithm 1 from \cite{li2024derivative} and utilize the posterior mean approximation approach (SVDD-MP) to compute the soft value function. Because our objective favors lower scores, we multiply the scoring function by $-1$ to correctly assign higher weights to these preferred candidates.

\begin{figure}[ht]
    \centering
    \includegraphics[width=\linewidth]{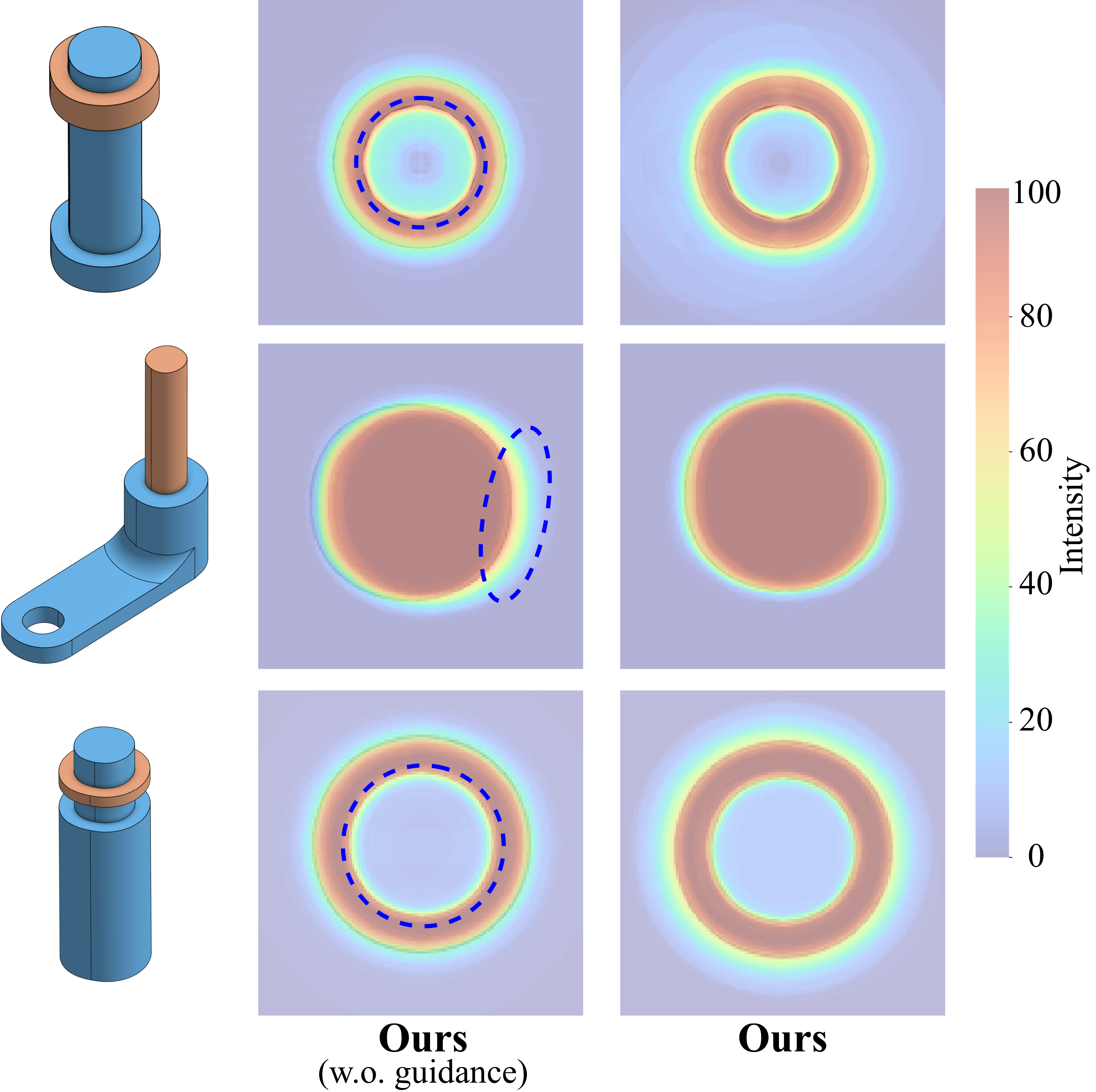}
    \caption{\textbf{Heatmap of Generated Shapes.} Visualization of the top-down view of $100$ generated CAD models using our method, with the \textcolor{blue}{blue} dashed ellipses highlighting the intersection areas with higher intensity.
    }
    \label{fig:sup-guidance-analysis}
\end{figure}

\section{Additional Visualization \& Results}
\label{sec:sup-additional-qualitative}

\subsection{Additional Guidance Analysis}
We provide more analysis of the guidance in Fig.~\ref{fig:sup-guidance-analysis}. In the first and third examples, the rings generated by our method with guidance have a more accurate inner diameter and a larger overall size compared to the generations without guidance. It is likely due to the semantic preservation term encourages the model to preserve the global shape ratio, while the geometry alignment term pushes the inner diameter to be large enough to fit the condition contact surfaces.

\subsection{Robustness Analysis}
\noindent \textbf{How robustness does our propose methods perform in noisy scenarios?} We evaluate methods on a noisy test set which is created by randomly adding, removing one contact face, or replacing with other faces on the test set. Table~\ref{table:noisy-eval} reports the results. Since our cost functions avoid collapsing optimized faces to the same shape as the condition faces, a small proportion of noisy input contact faces might not affect the overall guiding cues.

\begin{table}[h]
    \centering
    \renewcommand{\tabcolsep}{6.5pt}
    \resizebox{0.9\linewidth}{!}{
        \begin{tabular}{r | cccc}
            \toprule
            \textbf{Method} & $\textbf{CD} \downarrow$ & $\textbf{PR} \downarrow$ & $\textbf{IV} \downarrow$ & $\textbf{VR} \uparrow$ \\ 
            \midrule
            PivotMesh~\cite{weng2024pivotmesh} & 144.42 & - & - & - \\
            BrepGen~\cite{xu2024brepgen} & 119.23 & 0.47 & 19.11 & \textbf{0.46} \\
            MatchMaker~\cite{wang2025matchmaker} & 118.01 & 0.40 & 18.37 & 0.09 \\
            \midrule
            Ours (w.o. guidance) & 108.79 & 0.36 & 16.48 & 0.35 \\
            \rowcolor[HTML]{FFF2CC} Ours & \textbf{100.50} & \textbf{0.32} & \textbf{14.15} & 0.34 \\
            \bottomrule
        \end{tabular}
    }
    \caption{\textbf{Noisy contact scenarios.} \label{table:noisy-eval}}
\end{table}

\begin{table}[h]
    \centering
    \renewcommand{\tabcolsep}{6.5pt}
    \resizebox{0.9\linewidth}{!}{
        \begin{tabular}{r | cccc}
            \toprule
            \textbf{Method} & $\textbf{CD} \downarrow$ & $\textbf{PR} \downarrow$ & $\textbf{IV} \downarrow$ & $\textbf{VR} \uparrow$ \\ 
            \midrule
            PivotMesh~\cite{weng2024pivotmesh} & 151.35 & - & - & - \\
            BrepGen~\cite{xu2024brepgen} & 114.23 & 0.45 & 15.11 & 0.48 \\
            MatchMaker~\cite{wang2025matchmaker} & 112.08 & 0.51 & 27.83 & 0.13 \\
            \midrule
            Ours (w.o. guidance) & 98.45 & \textbf{0.17} & 9.82 & \textbf{0.53} \\
            Ours (w.o. $D_\text{sem}$) & \textbf{96.18} & 0.18 & \textbf{6.95} & 0.51 \\
            \rowcolor[HTML]{FFF2CC} Ours & 100.76 & 0.18 & 8.10 & 0.52 \\
            \bottomrule
        \end{tabular}
    }
    \caption{\textbf{Complex contact scenarios.} \label{table:complex-contact}}
\end{table}

\noindent \textbf{How does the scoring function perform in more complex cases?} We define ``more complex'' cases as instances where two assembled CAD parts share no equal boundary edge lengths. In these scenarios, our assumption of edge-length equality no longer holds. Although the test set already includes these cases, we isolate them to specifically analyze the behavior of our scoring function. As reported in Table~\ref{table:complex-contact}, our proposed network outperforms the baselines. Furthermore, our model equipped with the proposed guiding cue maintains competitive performance even under these challenging scenarios. While the variant that excludes the Semantic Preservation term (w.o. $D_\text{sem}$) achieves the best CD and IV, it produces the least watertight CAD models (the lowest VR) among our three method variants. Intuitively, this low VR can be attributed to the inconsistency between the guiding cues and the diffusion trajectory. In these complex scenarios, the predicted guiding cues are less informative or incorrect. These cues cause a significant discrepancy between the guidance direction and the diffusion path, leading to an unexpected distribution and the generation of invalid CAD geometries.

\subsection{Dataset Visualization}
\begin{figure*}[ht!]
    \centering
    \includegraphics[width=1.04\linewidth]{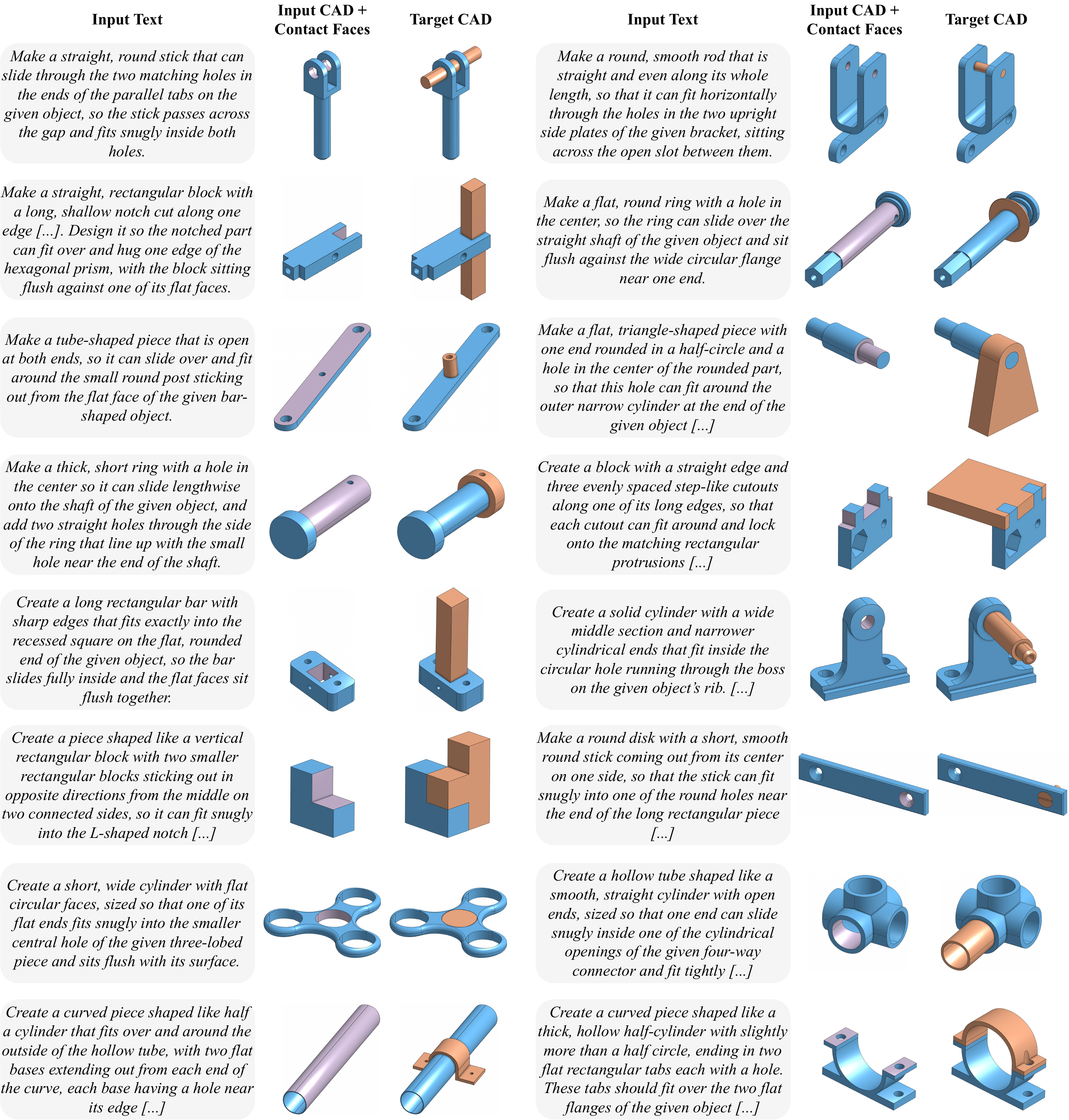}
    \caption{\textbf{Examples from KnitCAD dataset.} We demonstrate examples from KnitCAD. The condition CAD models are in \textcolor{lightblue}{\textbf{blue}}, the target CAD models are in \textcolor{lightorange}{\textbf{orange}}, and the desired contact faces are in \textcolor{lightpurple}{\textbf{purple}}.}
    \label{fig:sup-dataset-demo}
\end{figure*}

Fig.~\ref{fig:sup-dataset-demo} shows examples from our KnitCAD dataset. 

\subsection{Additional Qualitative Results}
\begin{figure*}[ht]
    \centering
    \includegraphics[width=\linewidth]{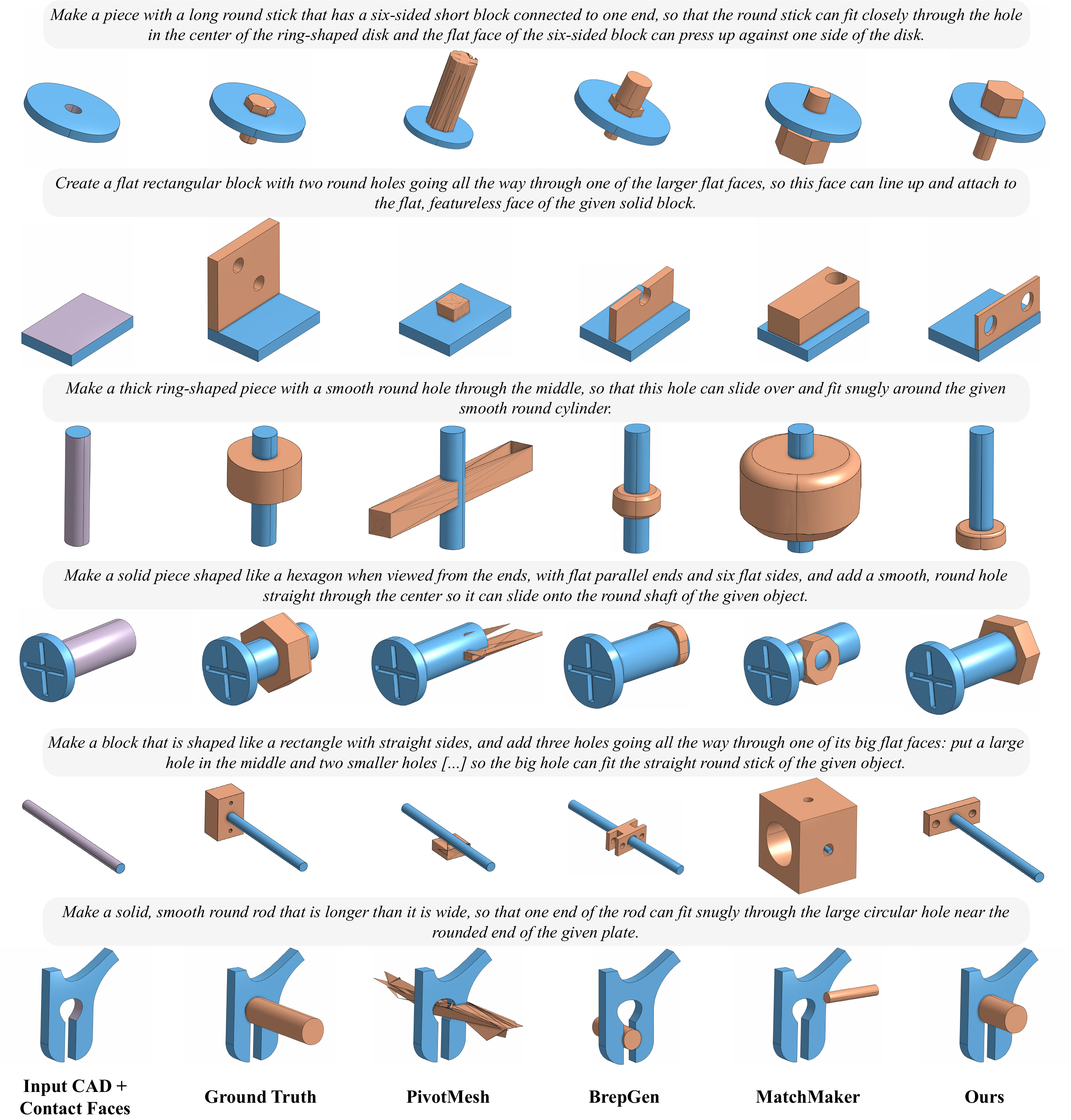}
    \caption{\textbf{Qualitative Results.} We demonstrate the qualitative results of our method and the other three baselines. The condition CAD models are in \textcolor{lightblue}{\textbf{blue}}, the generated CAD models are in \textcolor{lightorange}{\textbf{orange}}, and the desired contact face are in \textcolor{lightpurple}{\textbf{purple}}.}
    \label{fig:sup-qualitative-results}
\end{figure*}

We also show additional qualitative results from our method and the baselines in Fig.~\ref{fig:sup-qualitative-results}. Since MatchMaker relies on CAD autocompletion, it often generates objects that are larger than desired. In contrast, our method applies guidance at selected steps, which helps the generated objects remain semantically aligned with the text while also fitting the condition objects geometrically. 

\section{Demonstration of Practical Application}
\label{sec:sup-practical-application}

We present a preliminary study to demonstrate how our model supports interactive, 
sequential part generation in a real-world design workflow. Specifically, a human engineer and our model collaboratively assemble a label roll holder in a sequential workflow, where at each step the engineer selects contact faces on the existing assembly and provides a natural language prompt, and the model generates a new part.

\paragraph{Workflow.}
The assembly proceeds in five steps. The engineer begins by manually creating a base with a rectangular slot. Our model then generates an upright bracket conditioned on the slot geometry. The engineer duplicates and positions the bracket to form a symmetric two-upright configuration. Next, the model generates a cylindrical axle that threads through the aligned holes of both uprights. Finally, a roller sleeve is generated to fit over the axle, completing the assembly. The prompts used at each model-assisted generation step are shown below.

\begin{tcolorbox}[title=Prompt synthesizing ``upright bracket'']
\small
Create a flat, rectangular bar that fits snugly into the slot, with a centered circular 
through-hole positioned close to the top edge.
\end{tcolorbox}

\vspace{0.1em}

\begin{tcolorbox}[title=Prompt synthesizing ``axle rod'']
\small
Create a long cylindrical rod that threads through the vertical bar's hole and protrudes 
from both sides.
\end{tcolorbox}

\vspace{0.1em}

\begin{tcolorbox}[title=Prompt synthesizing ``roller sleeve'']
\small
Create a thick roller-shaped cylinder that covers most of the rod's length, featuring a 
central through-hole for a tight fit on the rod.
\end{tcolorbox}

\end{document}